\documentclass[Afour,sageh,times]{sagej}

\usepackage{moreverb,url,bm, mathrsfs,amssymb}
\usepackage{pifont}
\usepackage{tabularx, threeparttable}
\usepackage{multirow}
\usepackage{makecell}

\newcommand{\cmark}{\ding{51}}%

\newcommand{\xmark}{\ding{55}}%

\usepackage{subfigure}
\usepackage{xspace}
\usepackage[colorlinks,bookmarksopen,bookmarksnumbered,citecolor=red,urlcolor=red]{hyperref}
\usepackage{xcolor}
\newcommand\BibTeX{{\rmfamily B\kern-.05em \textsc{i\kern-.025em b}\kern-.08em
T\kern-.1667em\lower.7ex\hbox{E}\kern-.125emX}}
\newcommand{\regnet}{REGNetV2\xspace}
\def\volumeyear{2016}

\begin{document}


\title{REGNet V2: End-to-End REgion-based Grasp Detection Network for Grippers of Different Sizes in Point Clouds}
\author{Binglei Zhao, Han Wang, Jian Tang, Chengzhong Ma, Hanbo Zhang, Jiayuan Zhang, Xuguang Lan, and Xingyu Chen}


\affiliation{Xi’an Jiaotong University, China}
\corrauth{Xuguang Lan, Institute of Artificial Intelligence and Robotics, the National Engineering Laboratory for Visual Information Processing and Applications, School of Electronic and Information Engineering, Xi’an Jiaotong University, No.28 Xianning Road, Xi’an, Shaanxi, China.}
\email{xglan@mail.xjtu.edu.cn}

\begin{abstract}
Grasping has been a crucial but challenging problem in robotics for many years. One of the most important challenges is how to make grasping generalizable and robust to novel objects as well as grippers in unstructured environments. We present \regnet, a robotic grasping system that can adapt to different parallel jaws to grasp diversified objects. To support different grippers, \regnet embeds the gripper parameters into point clouds, based on which it predicts suitable grasp configurations. It includes three components: Score Network (SN), Grasp Region Network (GRN), and Refine Network (RN). In the first stage, SN is used to filter suitable points for grasping by grasp confidence scores. In the second stage, based on the selected points, GRN generates a set of grasp proposals. Finally, RN refines the grasp proposals for more accurate and robust predictions. We devise an analytic policy to choose the optimal grasp to be executed from the predicted grasp set. To train \regnet, we construct a large-scale grasp dataset containing collision-free grasp configurations using different parallel-jaw grippers. The experimental results demonstrate that \regnet with the analytic policy achieves the highest success rate of $74.98\%$ in real-world clutter scenes with $20$ objects, significantly outperforming several state-of-the-art methods, including GPD, PointNetGPD, and S4G. 
The code and dataset are available at https://github.com/zhaobinglei/REGNet-V2.
\end{abstract}


\maketitle

\section{Introduction}
In the real world, reliable robotic grasping is a crucial but challenging task. It plays a pivotal and fundamental role in robotic manipulation and interaction. Many factors affect grasp detection performance, such as unstructured environments, sensor noise, various object geometries, and gripper (end-effector) attributes. Grasping strategies designed for a single specific gripper are difficult to generalize to novel grippers, potentially leading to a large drop in performance. However, it is time-consuming to extensively re-collect data and re-train the policy for novel grippers. Since various grippers have their advantages and limitations when dealing with different tasks in robotic grasping and manipulation, it is inevitable to design a grasp policy suitable for arbitrary grippers. 

\begin{figure}[thbp]
\centering
\subfigure[]{
\begin{minipage}[c]{0.27\linewidth}
\centering
\includegraphics[height=2.8cm,width=2.3cm]{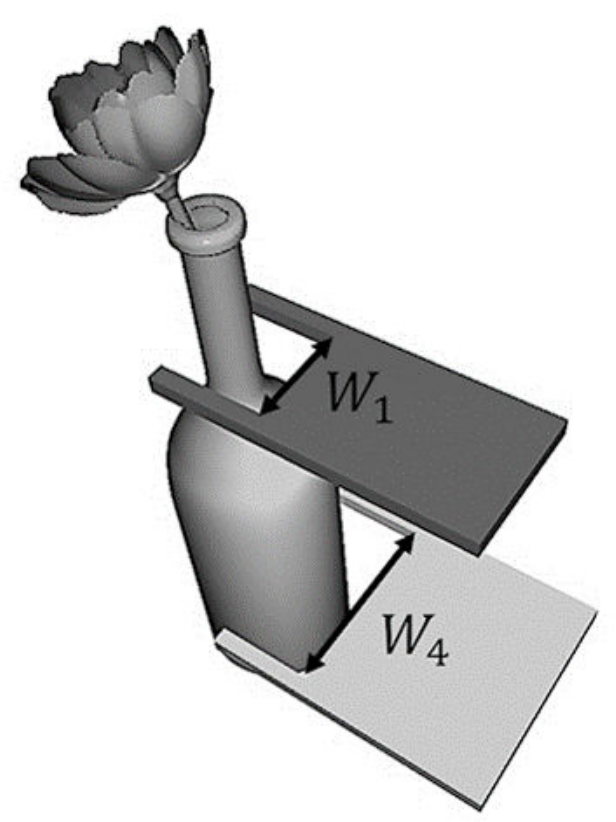}
\end{minipage}%
}
\subfigure[]{
\begin{minipage}[c]{0.38\linewidth}
\centering
\includegraphics[height=2.8cm,width=2.1cm]{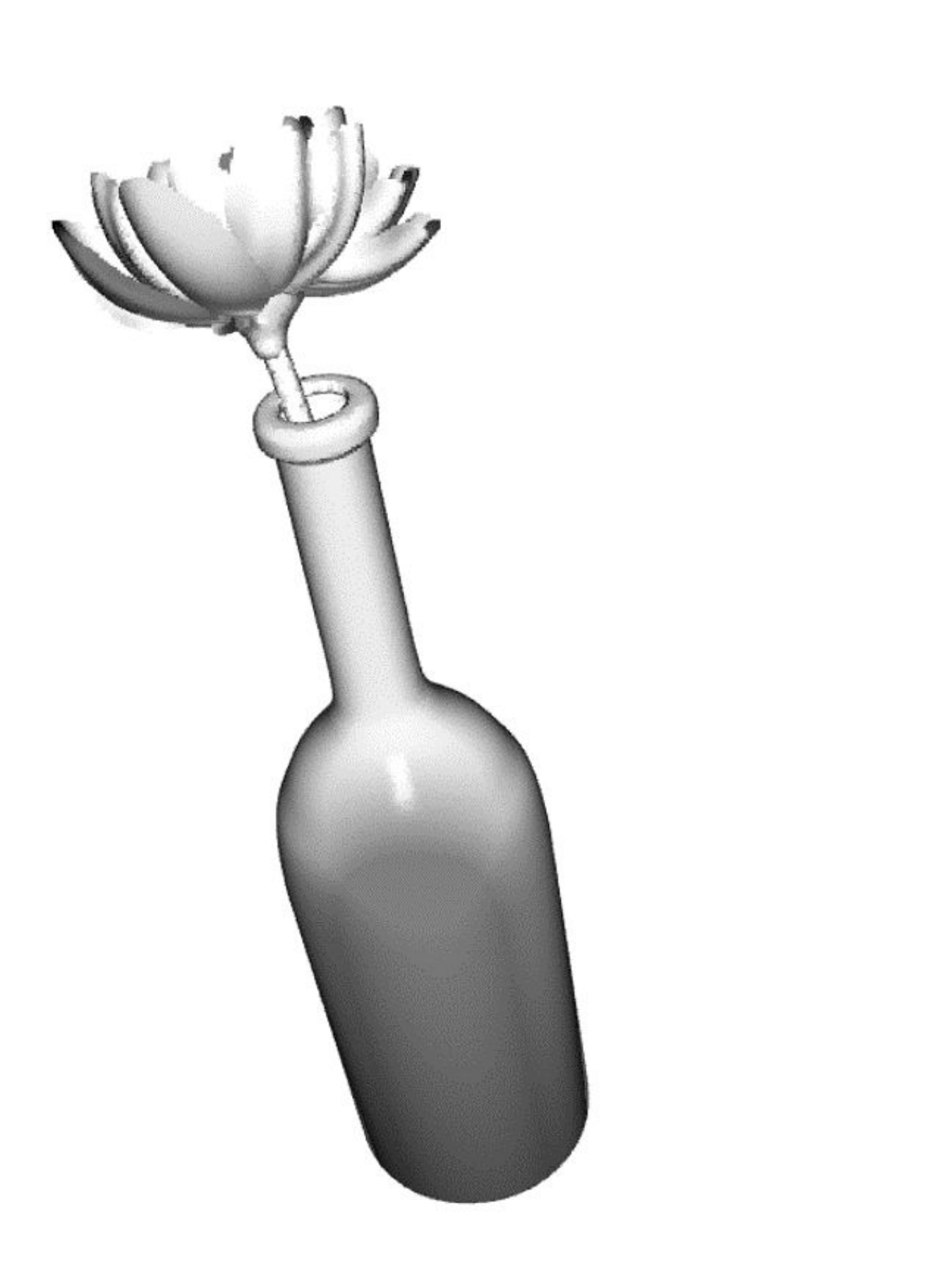}
\end{minipage}%
}
\subfigure[]{
\begin{minipage}[c]{0.23\linewidth}
\centering
\includegraphics[height=2.8cm,width=2.1cm]{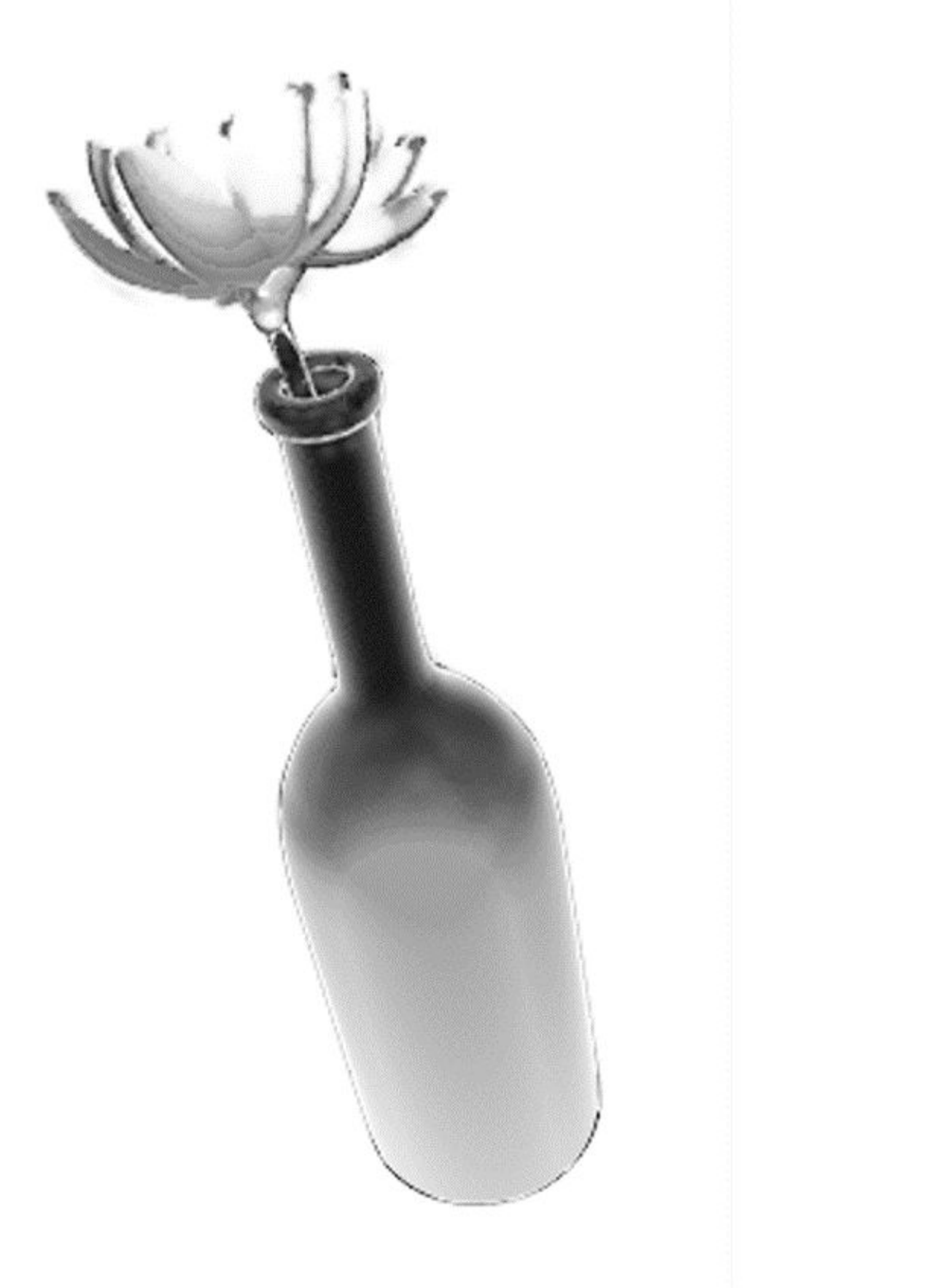}
\end{minipage}%
}
\caption{(a) Different suitable grasps generated for grippers with different sizes.
When grasping a bottle, 
(b) For a gripper with a bigger width, points closer to the bottom of the bottle have a higher grasp confidence.
(c) For a gripper with a smaller width, points closer to the bottleneck have the higher grasp confidence.}
\label{fig:diff_width}
\end{figure}

The capability of generalizing to various objects and grippers in unstructured environments would benefit a robot manipulation system. It requires the grasping algorithm not only to adapt to complex environments and novel objects but also to have the ability to generate stable grasps with high-quality scores for various grippers. As illustrated in Figure~\ref{fig:diff_width}(a), a narrower gripper is difficult to grasp larger objects without collision. Suitable grasps are different for various grippers when grasping the same object.

Traditional analytic methods (\cite{force_closure2,1992_planning_optimal_grasp,1995_3finger,1997_4finger,1999_parallel-jaw}) are difficult to generate grasps for novel object models and gripper models. Compared with those traditional solutions, learning-based methods trained on large amounts of labeled data show superiority in the adaptability of detecting grasps on novel objects.

Methods trained with manually labeled datasets (\cite{2011_rectangle,2015_deep,2015_real_convolutional,2018_anchor,2019_roi}) use algorithms akin to object detection to predict grasps. They struggle to find the optimal grasp position due to ignoring physical analysis and grasp quality metrics when grasping. Since the representation of the point cloud can naturally fuse depth with RGB information and handle observations from different perspectives through rotation and normalization operations, recent works (\cite{2019_pointnetgpd,2020_s4g,2020_pointnet++_grasping}) generate grasps based on the point cloud. However, these methods are trained on a single specific gripper, which cannot maintain consistent performance when generalizing to novel grippers.

\begin{figure}
\centering
\includegraphics[height=3.7cm,width=8.5cm]{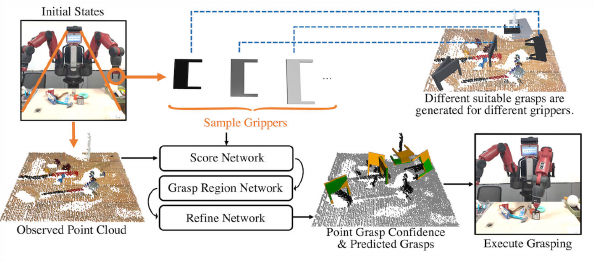}
\caption{The REGNetV2 pipeline. 
SN embeds the sampled gripper parameters to the observed point cloud and utilizes them to predict the point grasp confidence. 
Then GRN predicts grasp proposals based on predicted grasp confidence. 
After RN refines the proposals, we select the optimal grasp through the analytic grasp selection strategy to execute grasping.}
\label{fig:pipline}
\end{figure}

This article proposes an end-to-end gripper-embedded grasp network, which aims to produce suitable grasp configurations for grippers with different geometries on various objects, even novel objects in unstructured environments. Our system explicitly encodes sampled gripper parameters for the input to keep the corresponding physical meaning and embeds the representations into the observed point cloud. It predicts lots of grasps with corresponding grasp quality scores. Among the set of predicted grasps, we design an analytic grasp selection strategy by quantitatively analyzing the respective effectiveness of various scores on grasping performance. The pipeline of \regnet devised in our system consists of three stages: Score Network (SN), Grasp Region Network (GRN), and Refine Network (RN), which is indicated in Figure~\ref{fig:pipline}.

(A) Score Network: Taking the fused observation as input, for each point in the observed scene, the SN generates the \textbf{point grasp confidence} to record the ratio of successful grasp in a region around a given point. For grippers with different parameters, it will select different suitable grasp positions (\textbf{positive points} with high grasp confidence) for further grasp detection.

(B) Grasp Region Network: Different from anchor-free methods that use single-point features (\cite{2020_s4g,2020_pointnet++_grasping}), we introduce the spherical \textbf{grasp region} centered on the selected positive points from RN and \textbf{grasp anchor mechanism} predefining grasp orientations to more accurately predict grasp proposals in GRN. Compared to a single point, the grasp region contains surrounding points and provides a larger receptive field, enabling efficient local feature aggregation.

(C) Refine Network: To generate more precise grasps, RN fuses the features of grasp regions and the areas within the proposals to conduct proposal refinement, fully utilizing all information provided by SN and GRN. The point grasp confidence and predicted grasp configurations will be dissimilar for grippers with different parameters. After generating a set of grasps, we propose a novel analytic grasp selection strategy to choose the optimal grasp for executing.

To further train \regnet, we construct a simulated synthetic grasp dataset containing grasp configurations suitable for parallel-jaw grippers with different parameters. Our training dataset contains 4794 clutter scenes and 19176 partial point clouds observed from 4 different perspectives. Each observed point cloud provides the corresponding point grasp confidence and about $1000$ collision-free grasps with quality scores for different grippers. Experiments show that \regnet outperforms state-of-the-art grasping algorithms on diverse parallel jaws and can also generalize to grippers with novel parameters. To summarize, our contributions are as follows:

\begin{enumerate}
\item[(1)] We design an exquisite end-to-end grasp prediction network that combines a simple structure to effectively predict grasps for different parallel jaws from partial noisy point clouds. Our method outperforms several state-of-the-art point-cloud-based algorithms.
\item[(2)] Through quantitative analysis of the grasp quality score and vertical score, we devise an analytic strategy to select the optimal grasp to be executed from the predicted grasps, which improves the grasping success rate in actual robotic experiments.
\item[(3)] We automatically construct a large-scale multi-gripper grasp dataset in simulation, which can be used to train data-driven networks and improve the performance of grasp detection for different grippers.
\item[(4)] We design various real-world experiments, including grasping in single-object scenes, bin-picking, and grasping under different views. These experiments demonstrate that our system can successfully grasp novel objects in unstructured scenarios and adapt well to parallel jaws with different parameters.
\end{enumerate}

\section{Related Work}
Existing grasp detection methods are generally divided into two categories: analytic and empirical approaches (\cite{grasp_overview}). 
The analytic works generate grasps on given models using physical analysis methods, such as force-closure analysis (\cite{force_closure1, force_closure2})and grasp wrench space (GWS)  
 (\cite{gws1,gws2,fc_gws}).
These traditional methods can detect grasps for arbitrary grippers once they take into account the gripper model. However, they struggle to generate grasps for generic objects without 3D models, while empirical approaches based on data-driven methods can conveniently transfer to novel objects. In the following, we mainly discuss the learning-based methods. The vast majority of the research learns grasp configurations from input object features extracted from 2D and 3D observations, as well as other modalities (\cite{data_driven_survey}).

\subsection{Specific-Gripper Grasp Detection Method}

Most learning-based grasp algorithms can be divided into sampling-based and end-to-end methods. Sampling-based methods typically consist of a grasp candidate sampler and a grasp quality evaluator.

For image inputs, some methods choose grasping points and then predict the quality of sampled grasps (\cite{1996_visual_grasp}). For depth images with gradients, Dex-Net (\cite{2016_dexnet1,2017_dexnet2}) uses geometric heuristics to sample antipodal grasp candidates and ranks them with a CNN classifier.

The representation of the point cloud is convenient for describing the 3D geometric shape of an object and estimating the corresponding surface normal and Darboux frame for each point. Therefore, some methods (\cite{2016_gpd,2017_gpd_ijrr,2019_pointnetgpd,2019_sim2real}) sample candidates and assess the grasp quality based on the point cloud. However, when the input points are sparse and noisy, accurately calculating the Darboux frame can be challenging, resulting in inefficient candidate sampling (\cite{2020_s4g}).

Other approaches, such as \cite{2019_6dof_graspnet} and \cite{2019_6dof_target_driven}, use trained variational autoencoders to sample grasps instead of heuristic sampling. However, repeatedly scanning candidates and classifying them can be time-consuming.


End-to-end methods take the input raw observations and directly predict the grasp to be executed, avoiding the time consumption of repeated scanning and evaluating candidates. For image inputs, \cite{2008_grasp_point} directly identifies good grasp points from images.

With the development of deep learning theory and hardware, researchers have designed complex grasp networks to learn rotated rectangle representations instead of low-dimensional point representations (\cite{2011_rectangle,2015_deep,2015_real_convolutional,2016_dcnn,2017_real_time_grasp}).

\cite{2018_multigrasp} discretizes the grasp orientation and introduces oriented anchors to predict grasps following the Faster R-CNN framework (\cite{2015faster_rcnn}). Building upon the oriented anchor box, \cite{2018_anchor}, \cite{2019_oriented_anchor}, and \cite{2019_roi} propose efficient anchor matching strategies and ROI-based strategies to further improve grasp detection performance.

The grasp detection problem can also be formulated as a segmentation problem. Some researchers predict pixel-level grasp maps, similar to heatmaps in segmentation algorithms, to determine where to grasp (\cite{2018_graspnet_low_eff,2018_close,2019_DSGD,2021_gknet,2021_graspme}). Others utilize rectangular representations to predict grasps but refine the predicted grasp candidates by incorporating pixel-wise semantic segmentation (\cite{2021_end_trainable}).

With the proliferation of 3D vision-based techniques (\cite{2017_pointnet,2017_pointnet++,2018_sonet,2018_voxelnet,2019_hough_voting,2019_pointconv,2019_point_voxel,2021_point_transformer}), some works focus on predicting grasps with point clouds as input. Some researchers (\cite{2020_s4g,2020_pointnet++_grasping,2021_contact_graspnet}) design one-stage networks to regress grasps directly from features extracted by PointNet++. Other multi-stage grasp detection methods propose anchor-based mechanisms (\cite{2020_grasp_proposal_net}) and consider the geometry awareness of the local grasp region (\cite{2021_regnet,2021_gpr,2021_graspness}). Some algorithms design reachability (\cite{2020_reachability}), collision (\cite{2021_simultaneous}), and score prediction networks (\cite{edgegrasp}) to ensure that the generated grasps can be reached without collision. However, these methods are trained for a particular end-effector and do not consider the influence of the gripper shape on the stability of predicted grasps.

Except for point cloud-based scene representation, voxel-based scene representation (\cite{VGN} and \cite{VPN}) is also a popular algorithm for grasp detection. These methods extract the point cloud from the Truncated Signed Distance Function (TSDF) as the center of grasp and train a score prediction network. However, voxel-based grasp detection methods may struggle to accurately capture the details and shapes of object surfaces. Point cloud-based approaches offer higher precision and accuracy as they can capture the normals and curvature information of object surfaces.

\subsection{Multi-Gripper Grasp Detection Method}
For a novel gripper, its attributes such as geometry and kinematics can significantly affect the quality of the predicted grasp. However, generating grasp datasets and retraining networks for novel grippers can be time-consuming. Therefore, some algorithms (\cite{2020_unigrasp,2021_adagrasp}) encode gripper features into the observation.
UniGrasp (\cite{2020_unigrasp}) concatenates the extracted gripper features and object features in separate lower-dimensional latent spaces, while AdaGrasp (\cite{2021_adagrasp}) calculates the cross-convolution between the gripper and scene encodings. Both algorithms encode the gripper geometry into a low-dimensional representation through trained neural networks. In contrast, our method explicitly encodes the important gripper parameters with specific physical meaning.
Different gripper encoding and feature fusion methods can have an impact on the grasp detection performance.

\subsection{Grasp Dataset}
The methods that take RGB images as input are usually trained on the Cornell dataset and Jacquard dataset (\cite{2018_jacquard_dataset}). Additionally, \cite{2019_roi} manually labeled the VMRD dataset for grasping overlapping objects in multi-object scenes. Some automatically constructed datasets (\cite{2017_dexnet2,2019_pointnetgpd,2020_graspnet_1billion,2021_regnet,2022_regrad}) utilize random trials in simulation or physical analysis to generate binary-valued grasp success metrics and meticulous grasp quality metrics. However, some of these methods may not adapt well to different sizes of grippers, and some of the score estimation methods they use result in sparse scores.

For adaptive-gripper grasp detection, UniGrasp (\cite{2020_unigrasp}) provides annotated object tuples with various gripper contact grasp points. AdaGrasp (\cite{2021_adagrasp}) is trained end-to-end with self-supervised grasping trials labeled in simulation. However, their grasping posture is still limited to being straight up and down, which may not be flexible enough.

To train \regnet, we construct a large-scale grasp dataset suitable for parallel-jaw grippers with different widths. This dataset includes not only collision-free grasps but also the point grasp confidence of each point in the observed point clouds, which defines the ratio of successful grasp in a region around a given point.

This paper extends our earlier work \cite{2021_regnet}. We introduce a novel question generation method for disambiguation and provide additional experiments and analyses.

\section{Problem Statement} 
\subsection{Assumptions} 
Our model is based on the following assumptions:
(1) The geometry of the parallel-jaw gripper can be simplified as a regular geometry. This assumption allows us to model the gripper's shape and dimensions in a simplified manner.
(2) Each object to be grasped has a uniform mass distribution. This assumption simplifies the modeling of the object's physical properties and allows us to focus on the grasping strategy without considering variations in mass distribution.

\subsection{Definitions} 
Given an observed single-view point cloud, our goal is to generate appropriate parallel-jaw grasp configurations that are suitable for grippers of different sizes.

\begin{figure}[thbp]
\centering
\subfigure[]{
\begin{minipage}[t]{0.52\linewidth}
\centering
\includegraphics[height=3cm,width=4cm]{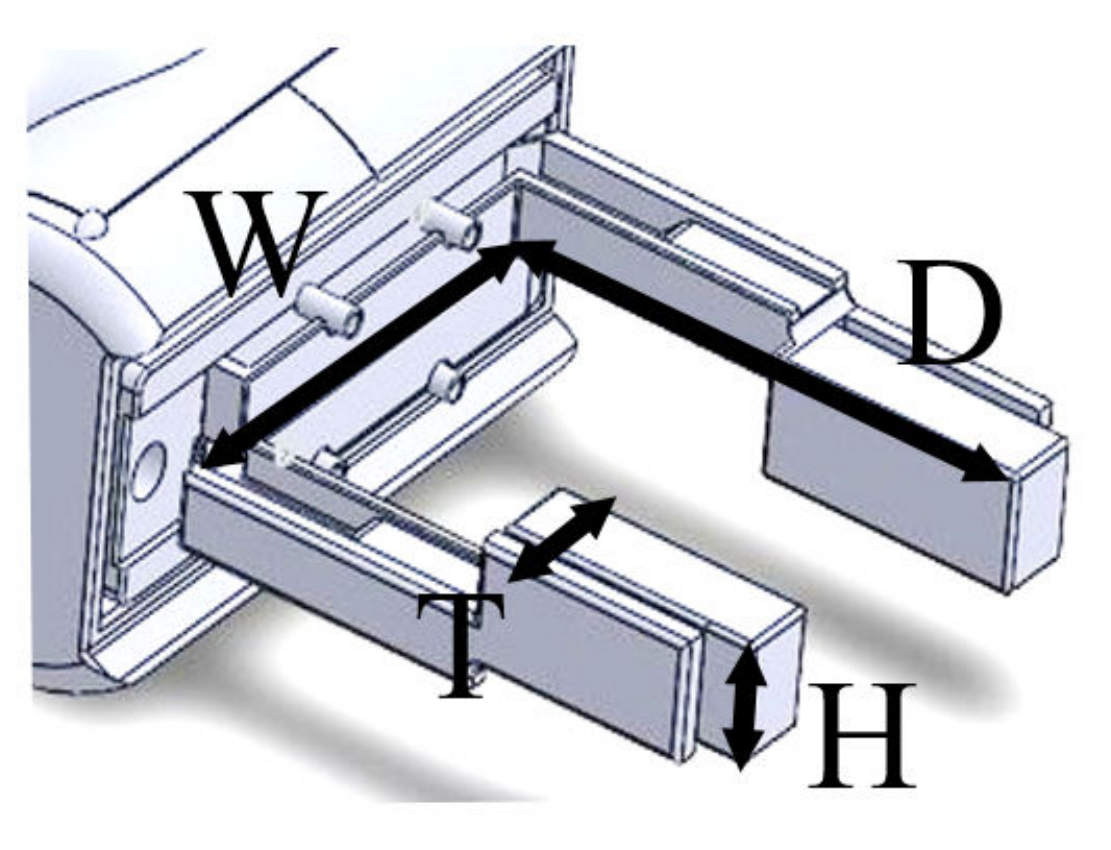}
\end{minipage}%
}%
\subfigure[]{
\begin{minipage}[t]{0.5\linewidth}
\centering
\includegraphics[height=3cm,width=4cm]{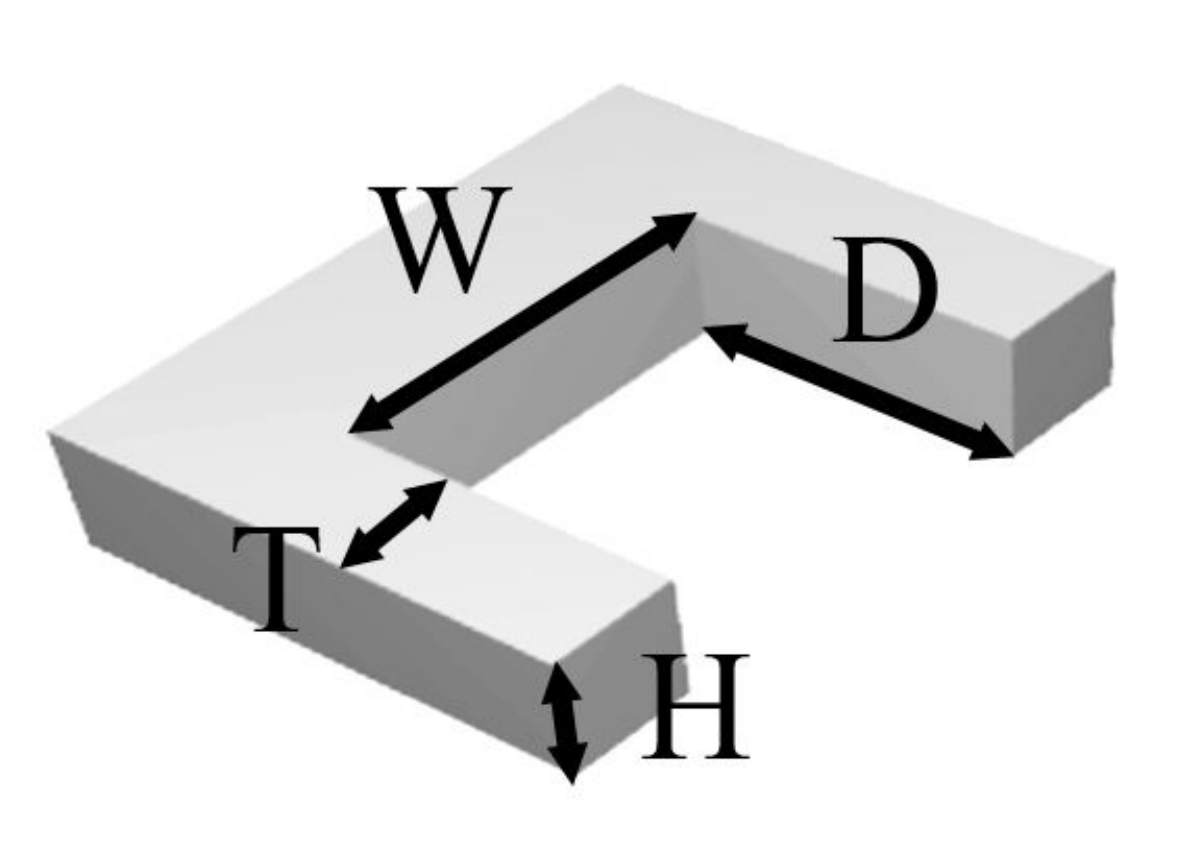}
\end{minipage}%
}%
\caption{(a) The visualization of a 2-finger parallel gripper.
(b) A simplified version of the gripper.
}
\label{fig:defination_gripper}
\end{figure}

\textbf{Point Cloud}: The observed single-view point cloud is represented by $\bm{V} \in \mathbb{R}^{N \times C}$, where $N$ is the number of points and $C$ is the number of features for each point. The features can include information such as location and color.

\textbf{Gripper Parameters}: The parameters of a 2-finger parallel gripper can be simplified as $S=(D, W, H, T)$, where $D$ represents the inner depth of the gripper, $W$ represents the inner width, $H$ represents the height, and $T$ represents the thickness of the gripper's fingers. The inner width of the gripper determines the maximum size of the object that can be grasped.

\textbf{Grasp}: A grasp is defined as $g=(c,\bm{r},\theta)\in\mathbb{R}^7$, following the definition used in PointNetGPD (\cite{2019_pointnetgpd}). Here, $c=(x,y,z)\in \mathbb{R}^3$ represents the grasp center, $\bm{r}=(r_x,r_y,r_z)\in \mathbb{R}^3$ represents the gripper orientation (Y-axis), and $\theta\in [-{\pi/2},{\pi/2}]$ represents the approach angle of the grasp.

\textbf{Grasp Quality Score}: The grasp quality score $s_q\in [0,1]$ is used to describe the probability of success for a grasp $g$. It is determined based on the force-closure property (\cite{force_closure1}) of the grasp, which measures the stability and effectiveness of the grasp.


\textbf{Grasp Vertical Score}: The grasp vertical score $s_v = 0.5 + \theta / \pi \in [0,1]$ represents the verticality of a grasp. It is calculated based on the approach angle $\theta$ of the grasp. A score of 0 indicates that the grasp is parallel to the X-Y plane, while a score of 1 indicates that the grasp is perpendicular to the X-Y plane. The vertical score is used to evaluate the alignment of the grasp with respect to the horizontal plane.

\textbf{Point Grasp Confidence}: The point grasp confidence $c_{p}\in [0,1]$ represents the suitability of a given point $p$ in the point cloud for grasping. It is defined as the ratio of successful grasps in a region around the point $p$ to assess which positions are suitable for grasping. The joint distribution $p(c_{p},g,s_q,\bm{V}, S)$ represents the relationship between the point grasp confidence, grasp, grasp quality scores, observed point cloud, and given gripper parameters.

\begin{figure}[thbp]
\centering
\subfigure[]{
\begin{minipage}[t]{0.45\linewidth}
\centering
\includegraphics[height=3cm,width=3.5cm]{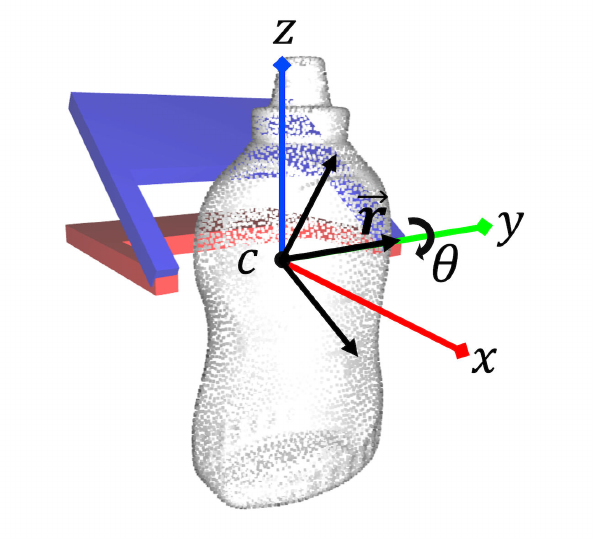}
\label{fig:defination_grasp}
\end{minipage}%
}%
\subfigure[]{
\begin{minipage}[t]{0.45\linewidth}
\centering
\includegraphics[height=3cm,width=3cm]{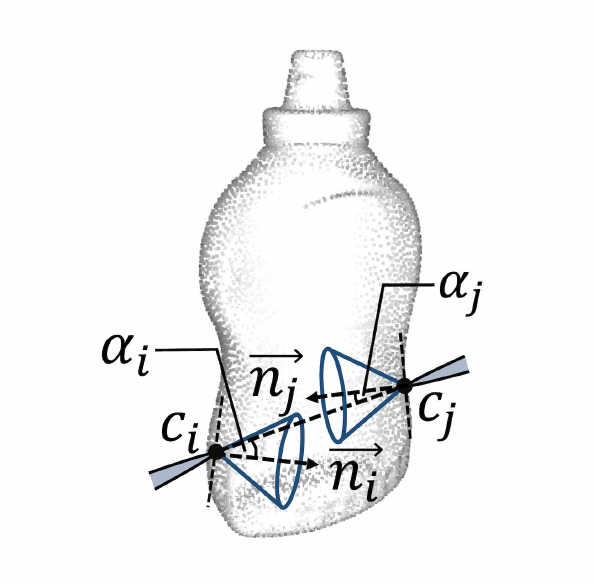}
\label{fig:defination_antipodal_score}
\end{minipage}%
}%
\caption{(a) {The definition of a grasp}. 
(b) When a gripper closes to grasp, two contacts $c_i, c_j$ are generated.
$\alpha_i, \alpha_j$ are angles between the force direction and normals $n_i, n_j$. Blue lines show friction cones (\cite{force_closure1}) generated at contacts.}
\end{figure}

\subsection{Objective} 
We denote the grasp detection model as $D(\bm{V}, S)$. Given an observed point cloud $\bm{V}$ and predefined gripper parameters $S$, our goal is to learn the robustness model $D_{\theta^*}(\bm{V}, S)$, which predicts the point grasp confidence $\hat{c_{p}}$, generates an appropriate grasp $\hat{g}$, and provides the corresponding quality score $\hat{s_q}$. The model aims to accurately estimate the suitability of each point for grasping, generate grasps that are suitable for the given gripper parameters, and provide a quality score that reflects the effectiveness and stability of the grasp.
$$
\theta^* = \mathop{\arg\min}_{\theta \in \Theta} \mathbb{E}_{p(c_{p},g,s_q,\bm{v},S)} [\mathcal{L}(t, D_{\theta}(\bm{V},S))]
$$
where $t=(c_{p},g,s_q)$ is the combined target of point grasp confidence, grasp, grasp quality score, $\mathcal{L}$ is the loss function defined in our proposed network \regnet, and $\Theta$ defines the parameters of \regnet.

\section{Multi-Gripper Dataset Construction}
To train \regnet, we construct a large-scale grasp dataset $G$ that is suitable for different gripper parameters. We vary the inner width $W$ of the gripper while keeping other parameters in $S$ consistent. The process of generating the dataset involves four steps:
(1) Simulating clutter scenes: We simulate cluttered scenes by randomly placing objects in a virtual environment. These objects can have different shapes, sizes, and orientations to create diverse and realistic scenarios.
(2) Constructing collision-free grasps: For each object in the clutter scene, we generate collision-free grasps $g$ using the gripper parameters $S$. These grasps are carefully selected to ensure that they can be executed without any collisions with the object or other obstacles in the scene.
(3) Computing grasp quality scores: We compute the corresponding grasp quality scores $s_q$ for each generated grasp $g$ based on the set gripper parameters $W$. The grasp quality scores reflect the stability and effectiveness of the grasp and are used as a measure of the grasp's quality.
(4) Generating observed point clouds and point grasp confidence: We generate observed point clouds $\bm{V}$ by simulating the depth camera's view of the clutter scene. Additionally, we compute the point grasp confidence $c_{p}$ for each point in the point cloud, which indicates the suitability of the point for grasping.
By following these steps, we construct a large-scale grasp dataset $G$ that contains a variety of clutter scenes, collision-free grasps, grasp quality scores, observed point clouds, and point grasp confidence. This dataset is used to train \regnet and improve its grasp detection performance for different gripper parameters.

\subsection{Simulating scenes}
Following the approach of S$^4$G \cite{2020_s4g}, we randomly select objects from the YCB dataset \cite{2017_ycb} and scale them by different factors. We use MuJoCo \cite{2012_mujoco} as our simulator and initialize a table with a height of $0.75m$. We load the selected object models into the scene with varying heights and random poses. We record the poses and positions of the objects when they are in equilibrium on the table plane. 
The point cloud of the scene is generated by combining the point clouds of the table and the objects. The simulated clutter scenes contain complete prior information, allowing us to generate collision-free grasps. This information is used during the grasp generation process to ensure that the generated grasps do not result in any collisions with the objects or the table.

\subsection{Constructing collision-free grasps with grasp quality scores and center scores} 
The construction of collision-free grasps involves two steps: 
(1) sampling a set of grasp candidates for each single object, 
(2) filtering out grasps that collide with the scene point cloud.
After obtaining the collision-free grasp, we compute the grasp quality score and the center score 
based on the force-closure analysis (\cite{force_closure1,force_closure2}) and the distance from the grasp to the object center, respectively.

\textbf{Object-specific grasps generation.} 
It is inefficient to search exhaustively in a clutter scene directly because of the repetitive search on the same object.
Therefore, we sample a set of grasp candidates for each single object instead of the whole scene.
To be specific, we randomly sample points on the object model and obtain the Darboux frames (\cite{2020_s4g}) centered at the selected points. 
Then we keep the grasp approaching vector (X-axis) the same as the normal of the selected point, 
and then sample the grasp candidates in the neighborhood of the Darboux frame. 

\textbf{Collision-free grasps construction.} 
According to the recorded poses and positions (6-DoF pose) of the objects, the object-specific candidates can be transformed into the scenes.
The set of the scene-specific candidates is the set of all transformed candidates in the scene.
We take different inner widths $W_i$ to compose different predefined gripper parameters $S_i$.
Then we perform collision detection respectively to each gripper $S_i$ and put the collision-free grasps into $G_i$.
Our dataset $G$ is the union of all $G_i$.

\textbf{Grasp quality scores generation.}
We assign the grasp quality score $s_q$ as the antipodal score, describing the force closure property (\cite{force_closure1}) of a grasp.
In Figure~\ref{fig:defination_antipodal_score}, each grasp has two contact points $c_i,c_j$ with the object.
We denote the angles between the force direction (connection direction of two contacts) and contacts' normals as $\alpha_i$ and $\alpha_j$.
We define the grasp quality score $s_q=cos\alpha_i \cdot cos\alpha_j$.
Regardless of friction coefficients, 
the larger the $ s_q $, the greater the possibility of a grasp to be force-closure.


\begin{figure*}
\centering
\includegraphics[height=8cm,width=17.2cm]{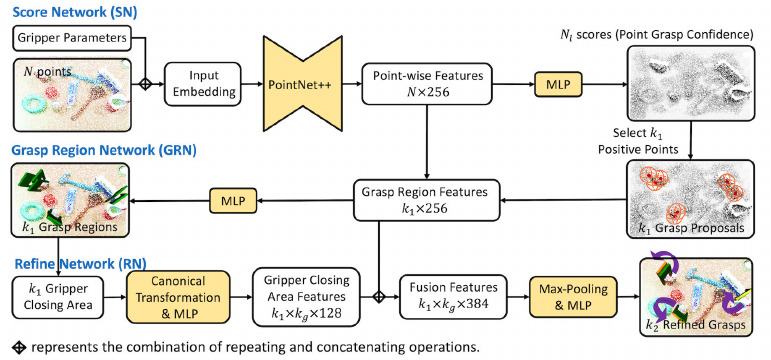}
\caption{The architecture of \regnet. SN takes the point cloud and the gripper parameters as input and outputs point grasp confidence. 
The darker point has a higher point grasp confidence.
GRN selects some points with higher confidence and predicts grasp proposals based on grasp regions centered on the selected points. 
RN then refines grasp proposals through the gripper closing area features and grasp region features.}
\label{fig:multi_architecture}
\end{figure*}

\subsection{Generating observed point clouds with point grasp confidence}
We utilize BlenSor (\cite{2011_blensor}) and Blender (\cite{2018_blender}) to render the observed point cloud $\bm{V}$.
We load the table and objects with the recorded poses and positions.
Then we load a Kinect with a known intrinsic matrix in different 6-DoF poses to collect different-view point clouds.
The single-view noisy point clouds are observed by the Kinect from different perspectives.

We assign the point grasp confidence $c_{p}$ describing 
the ratio of successful grasp in a region around the point $p \in \bm{V}$, which can help learn which position in $\bm{V}$ is suitable for grasping.
Noticeably, the grasps in $G_i$ and the corresponding point grasp confidence are generated for each different gripper $S_i$.
Specifically, we count all grasps in $G_i$ to calculate $c_{p}^i$ for the gripper $S_i$, which is defined as:
$$ 
c_{p}^i = tanh\sum_{g \in G_i} \sigma_{g} , \ p \in \bm{V}, \ i=1,2,3,4,  \eqno{(1)}
$$
\vspace{-0.28cm}
$$
\sigma_{g}=
   \begin{cases}
   0&\mbox{$dis(p, c_g) \ \textgreater \ d_{th}$}\\
   1- dis(p, c_g) / d_{th}&\mbox{else}
   \end{cases}  
$$
where $g$ is a grasp in $G_i$, 
$dis(p, c_g)$ is the distance between $p$ and the center of ${g}$,
and $d_{th}$ is the distance threshold.
If $dis(p, c_g) \textgreater d_{th}$, $g$ does not contribute to the grasp confidence of point $p$.
Intuitively, $c_{p}^i$ will be higher as there are more grasp annotations near $p$.
Therefore, the point grasp confidence is different for grippers $S_i$ with different sizes, which is illustrated in Figure~\ref{fig:diff_width}.

When simulating scenes, we select $131$ object models and randomly load $5-20$ models in the simulator. 
\subsection{Implementation Details}
\begin{table}[ht]
\small\sf\centering
\caption{Distribution of grasps generated for different grippers.}
\begin{tabular}{p{2.3cm}<{\centering}p{1.5cm}<{\centering}p{1.5cm}<{\centering}p{1.5cm}<{\centering}}
\toprule
\bf Gripper Width & \bf MNG & \bf MAS & \bf MCS   \\
\midrule
{0.06}  & 883  & 0.5648  &  0.8654\\
{0.08}  & 1373  & 0.5801  &  0.8665\\
{0.10}  & 2029  & 0.5472  &  0.8835\\
{0.12}  & 2450  & 0.5420  &  0.8914\\
\bottomrule
\end{tabular}
\label{table:distribution}
\end{table}
Then we randomly sample $2000$ points as the centers of sampled grasp candidates for single-object grasp generation.
When constructing collision-free grasps in each scene, we take the inner width $W$ from the set of $0.06$, $0.08$, $0.10$, $0.12m$ to get different predefined gripper parameters $S_i (i=1,2,3,4)$ and our dataset can be described as $G = \cup_{i=1}^{4} G_i$.
For observed point clouds generation, we predefine $4$ different 6-Dof poses for the Kinect, whose Euler angles and positions are ([0,0.6454,0], [0.8,0,1.7]), ([0,-0.2221,0], [-0.8,0,1.6]), ([0,0.6444,1.5708)], [0,0.75,1.7]) and ([0,0.7500,-1.5708], [0,-0.75,1.6]), respectively.
For each scene, we obtain $4$ different point clouds from $4$ different perspectives.
And $d_{th}$ is set as $0.01m$ for point grasp confidence calculation.

The training and validation dataset (Train/Val) totally contains $4794$ clutter scenes and $19176$ rendered single-view point clouds.
We split the dataset by a ratio of 4: 1 for training and validation.
The test dataset (Test) contains $757$ clutter scenes and $3028$ rendered point clouds observed from 4 different perspectives. 
The distribution of grasps and corresponding scores is illustrated in Table \ref{table:distribution}.
Following REGRAD (\cite{2022_regrad}), \textit{MNG} represents the mean number of grasps in a scene.
\textit{MAS} and \textit{MCS} denote the mean antipodal score and mean center score. 
For gripper with a larger width, it's easier to generate collision-free grasps.

\section{Method}
We present a gripper-embedded network, \regnet for 6-DoF grasp detecting. 
To generate grasps for grippers with different sizes, we explicitly embed the parameters of the gripper and observed point cloud to the inputs for grasp prediction.  
Since the gripper parameters have a specific physical meaning, we directly repeat them and concatenate them with the point cloud.
As illustrated in Figure~\ref{fig:multi_architecture}, the overall architecture includes three stages, Score Network (SN), Grasp Region Network (GRN), and Refine Network (RN).
The extracted features are shared with all three stages.
After predicting grasps, an analytic policy is devised to select the optimal grasp for executing.

\subsection{Score Network for Point Grasp Confidence Prediction}
Since gripper parameters influence the generation of point grasp confidence, 
The Score Network (SN) utilizes the observed point cloud $\bm{V}$ and the gripper parameters $S$ to estimate point grasp confidence $c_p$, the ratio of successful grasp in a region around a given point. 
To be more specific, the only different gripper parameter is the width $W$ when data generation, so we directly embed the width $W$ to the observation $\bm{V}$ as input.

\begin{figure}
\centering
\includegraphics[height=6cm,width=8.5cm]{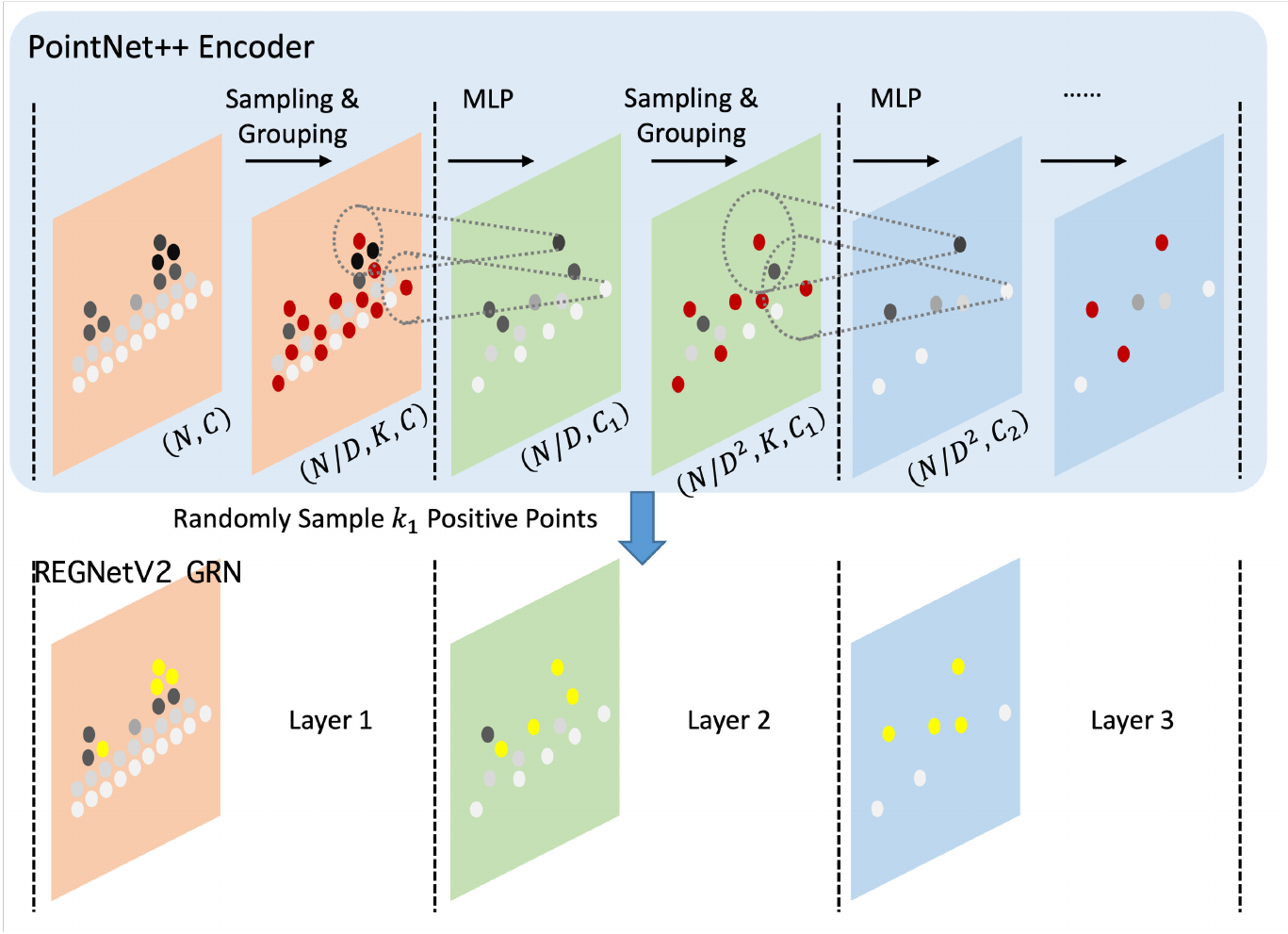}
\caption{Positive points selecting process.
In the PointNet++ encoder, group centers (red points) are sampled through the farthest point sampling method.
In GRN, we select the top $k_1$ points (yellow points) with high grasp confidence in different encoder layers.
}
\label{fig:fps_layer}
\end{figure}

To get the input embedding, we use the width $W$ as an additional feature of the point cloud by repeating and concatenating operations.
Specifically, the width $W$ is repeated to form a $N\times1$ tensor and concatenated to the $N\times C$ point cloud.
The backbone network, PointNet++ (\cite{2017_pointnet++}) encodes the input points with $C+1$ channels into group features and decodes these features into point-wise features through distance interpolation.
We do not add the width embedding to the group features extracted by the encoder but consider the impact of the gripper width on the features at the beginning of the encoding.
Then we use a multilayer perceptron (MLP) to regress point grasp confidence from the extracted features. 
The SN loss $L1$ is defined based on the MSE loss, formulated as (2).
$$
L_1 = \frac{1}{N}\sum\limits_{p \in \bm{V}} (c_p^i - \widehat{c}_p^i)^2, \ i=1,2,3,4, 
\eqno{(2)}
$$
where $N$ is the number of points, $c_p^i$ and $\widehat{c}_p^i$ are the ground-truth and predicted grasp confidence of point $p$.
The generation method of point grasp confidence is mentioned before.
We utilize all data generated with different gripper widths $(W_1-W_4)$ to train the Score Network.

\subsection{Grasp Region Network for Grasp Proposal Generation}

The point with high grasp confidence has a high probability of generating suitable grasps, which holds more valuable information for predicting associated grasp. 
Therefore, we select top $k_1$ positive points with high grasp confidence as the centers of regression.
Instead of single-point features, Grasp Region Network (GRN) uses grasp region features to regress grasp proposals at points with high grasp confidence.
Compared to a single point, the grasp region contains surrounding points and provides a larger receptive field, which effectively helps aggregate local features. 
Furthermore, we introduce the grasp anchor mechanism that predefines grasp anchors with assigned orientations to predict proposals more accurately.

Since increasing $k_1$ will significantly increase the processing time, we hope to increase the diversity of predicted grasps while keeping $k_1$ constant.
As illustrated in Figure~\ref{fig:fps_layer}, the deeper layer in the PointNet++ encoder has more advanced features and contains fewer points. 
Sampling positive points in deeper layers can cover a more diverse space, which means it is able to ensure the diversity of the sampled points in the spatial area under the premise of selecting higher confidence points.
After selecting $k_1$ positive points, we obtain corresponding grasp regions, and spheres centered on the positive points. 
Then we utilize the grasp region features to regress the grasp proposal based on the grasp anchor mechanism.

\textbf{Grasp region.} 
In the research of 2D object detection, “Region” is often considered to be a rectangle (\cite{2015faster_rcnn}). Nevertheless, the Grasp Region is a sphere centered on a positive point in this paper.

We use ball query (\cite{2017_pointnet++}) to find all points within a radius $\phi$ to a positive point $p_a$, which guarantees the grasp region has a fixed region scale. 
We randomly sample and keep $K$ points in each grasp region to ensure a fixed-dimensional input.
We concatenate the features of the points in a grasp region and then perform the max-pooling operation on them to get the \emph{Grasp Region Features}.
Then, we predict $k_1$ proposals based on the grasp anchor mechanism by applying MLP on the $k_1$ Grasp Region Features.

\begin{figure}
\centering
\includegraphics[height=4.2cm,width=8cm]{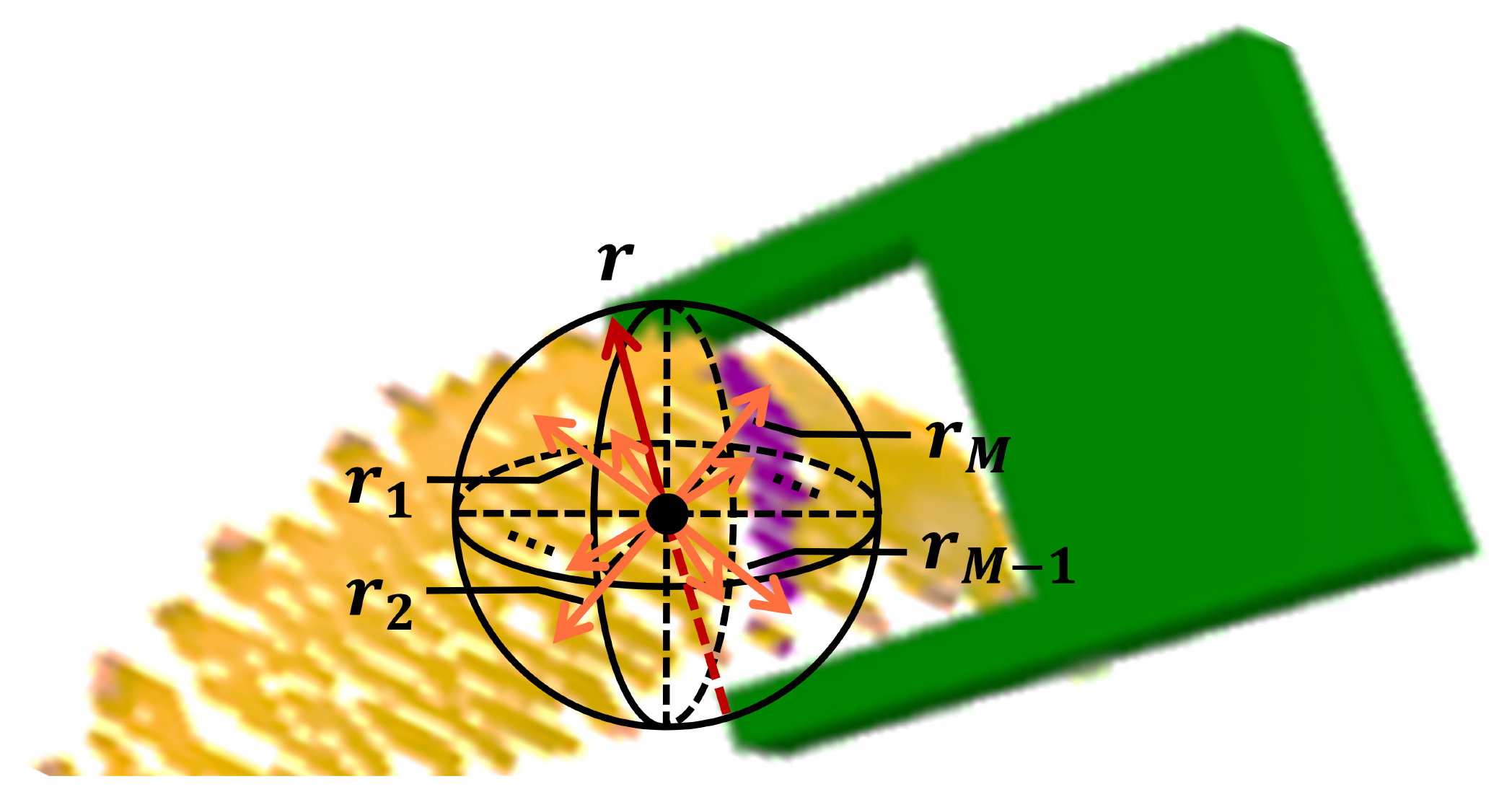}
\caption{Illustration of predefined assigned orientations. The red line shows
the ground-truth gripper orientation $\bm{r}$. The orientation category is determined by
the minimum angle between $\bm{r}$ and all assigned orientations $\bm{r}_i$ (orange lines).
}
\label{fig:anchor}
\end{figure}

\textbf{Grasp anchor mechanism.} 
As mentioned, a grasp is defined as $g=(c,\bm{r},\theta)$.
Since small changes in approach angle $\theta$ are marginal for grasp prediction, $\theta$ is permitted to have an acceptable regression error. 
Therefore, we perform anchor-based regression on gripper orientation $\bm{r}$ and simplify the regression of $\theta$ to direct regression.
We define the grasp anchor as $g_i = (p_p, \bm{r}_i, 0),(i = 1, 2, ..., M)$, where $p_p$ is the positive points, $\bm{r}_i$ is the assigned orientation, and M is the number of $\bm{r}_i$. 
Each anchor is centered at the positive point $p_p$ and is associated with a 3-dimensional orientation $\bm{r}_i$ and a zero approach angle as the reference. 
Since the grasp orientation can be any three-dimensional vector, without much prior knowledge, we only need to uniformly sample the assigned orientation $\bm{r}_i$ from the origin to a unit sphere surface, which is shown in Figure~\ref{fig:anchor}.

The orientation estimation loss consists of two terms, one for binary classification of $\bm{r}_i$, and the other for residual regression.
For each grasp anchor, its corresponding binary class label $o_i$ is defined as:
$$
o =
   \begin{cases}
   1  &\mbox{$ \langle \bm{r}, \bm{r}_i \rangle < \alpha_1$}\\
   0  &\mbox{$ \langle \bm{r}, \bm{r}_i \rangle \ge \alpha_2$}
   \end{cases}  , \ \ i \in \{1,2,...,M\}, 
$$
where $\bm{r}$ is the ground-truth gripper orientation. Note that the normalization of $\bm{r}$ and $\bm{r}_i$ makes them become the unit vectors.
When label generation, we use grasp anchors with discrimination for training.
And the regression residues to be optimized at $p_p$ are defined as follows. 
$$
{res}_r = \frac{\bm{r}}{\|\bm{r}\|} - \frac{\bm{r}_i}{\|\bm{r}_i\|}, \ {res}_c = (c-p_p)/c_b, 
$$
where ${res}_c,\ {res}_r$ are the ground-truth residues of the center, the orientation of each grasp anchor,
$c$ is the ground-truth grasp center,
and $c_b$ is a constant, used to balance the weight of grasp center regression loss.

We utilize smooth L1 loss to estimate grasp centers because the distance between $p_p$ and its corresponding grasp center is within a small range. 
We also adopt direct regression to estimate the approach angle $\theta$, and the grasp quality score $s_q$.
With these definitions, the overall GRN loss $L_2$ for all grasp anchors is defined as:
$$
L_2 = \frac{1}{k_1} \sum_{o \in \{0,1\} } \lambda_{cls} \cdot L_{cls}(o, \hat{o}) + \sum_{o=1} \sum_u \lambda_u \cdot L_{u}   ,  \eqno{(3)}
$$
$$
L_{u}=
   \begin{cases}
   SL(res_u, \hat{res_u})  &\mbox{$\ u \in \{ c, \bm{r} \}$}\\
   SL(u, \hat{u})          &\mbox{$\ u \in \{ \theta, s_q \}$}
   \end{cases}  ,
$$
where $L_{cls}, \ L_u$ are the orientation's classification loss and the residual regression losses, which denote the focal loss (\cite{2017_focal_loss}) and smooth L1 loss, respectively.
$o$ and $\hat{o}$ are the ground-truth and predicted category of orientation.
${res}_u, \hat{res}_u\ (u \in \{ c, \bm{r} \})$ are the ground-truth and predicted residues of center, orientation.
$u, \hat{u} \ (u \in \{ \theta, s_q \})$ are the ground-truth predicted angle and quality score. 
The predicted grasp proposal can be formulated as:
$$
\begin{aligned}
g_{GRN}&= (\hat{c}, \ \hat{\bm{r}}, \ \hat{\theta}) \\
&= (\hat{res}_c \cdot c_b + p_p, \ \bm{res}_r + \bm{r}_{\hat{o}}, \ \hat{\theta})
\end{aligned}
$$
Note that the predicted proposals only contain proposals with $\hat{o}=1$.
In practice, $c_b$ is set as $0.1$ to balance the grasp center regression loss.
We set $\gamma = 2.0, \alpha = 0.25$ in focal loss to address anchor class imbalance.
Considering the different magnitudes of the losses, we set $\lambda_{cls} = 0.2, \lambda_{u} = 1$ to balance the impact of each loss on $L_2$. 

\subsection{Refine Network for Grasp Refinement}
\begin{figure}
\centering
\includegraphics[height=5.4cm,width=8.4cm]{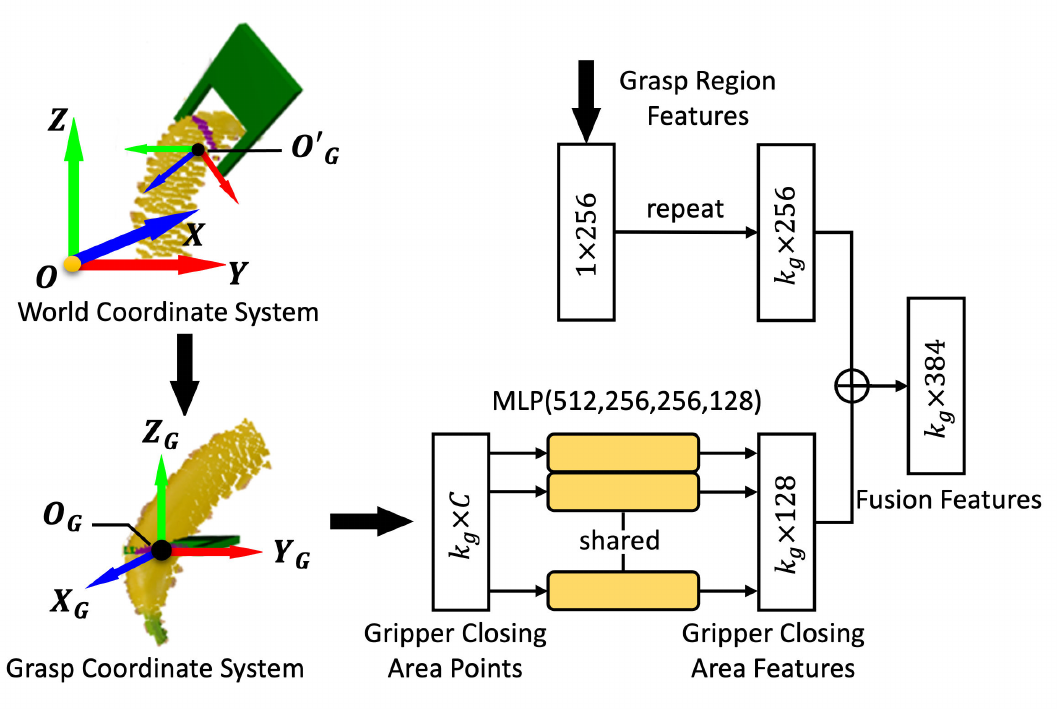}
\caption{Illustration of predefined assigned orientations. The red line shows
the ground-truth orientation $\bm{r}$. The orientation category is determined by
the minimum angle between $\bm{r}$ and all assigned orientations $\bm{r}_i$ (orange lines).
}
\label{fig:RN}
\end{figure}
Refine Network (RN) fully utilizes all information provided by SN and GRN to generate more accurate grasps. 
Since the grasp proposal predicted by GRN is closer to the ground truth, the area within the predicted grasp (\emph{gripper closing area}) contains information closer to the ground truth.
As shown in Figure~\ref{fig:RN}, RN uses both the gripper closing area (features) and grasp region (features) for grasp proposal refinement.
To be specific, we transform the points in the gripper closing area from the world coordinate system to the grasp coordinate system. 
Gripper closing area features are extracted from the transformed points through MLP.
RN then refines grasp proposals using fusion features of gripper closing areas and grasp regions.

\textbf{Canonical transformation.} 
For the predicted grasp $\hat{g} =(\hat{c}, \hat{\bm{r}}, \hat{\theta})$, we transform each point in the gripper closing area from the world coordinate system to grasp coordinate system.
As illustrated in Figure~\ref{fig:RN}, the grasp coordinate system defines that: 
(1) the origin $O_G$ is located at $\hat{c}$; 
(2) $Y_G$-axis is along $\hat{\bm{r}}$; 
(3) $X_G$-axis is the approach direction, which is obtained by rotating $X'$-axis around $Y_G$-axis by $\theta$, ($X'$ is parallel to the ground in the world coordinate system and perpendicular
to $Y_G$); 
(4) $Z_G$-axis is perpendicular to both $X_G$ and $Y_G$.

\textbf{Grasp refinement through fusion feature.}
As shown in Figure~\ref{fig:RN}, after obtaining transformed points in gripper closing areas, we apply MLP on these points to extract the local features, which are called \emph{gripper closing area features}. 
We replicate the grasp region features in GRN to each transformed point and concatenate them with the gripper closing area features.
RN further uses these fusion features to refine grasps through max-pooling and MLP operation.
This strategy fully utilizes the local shape information of grasps obtained from GRN.

\begin{figure*}[thbp]
\centering
\subfigure[]{
\centering
\includegraphics[height=4.5cm,width=5.2cm]{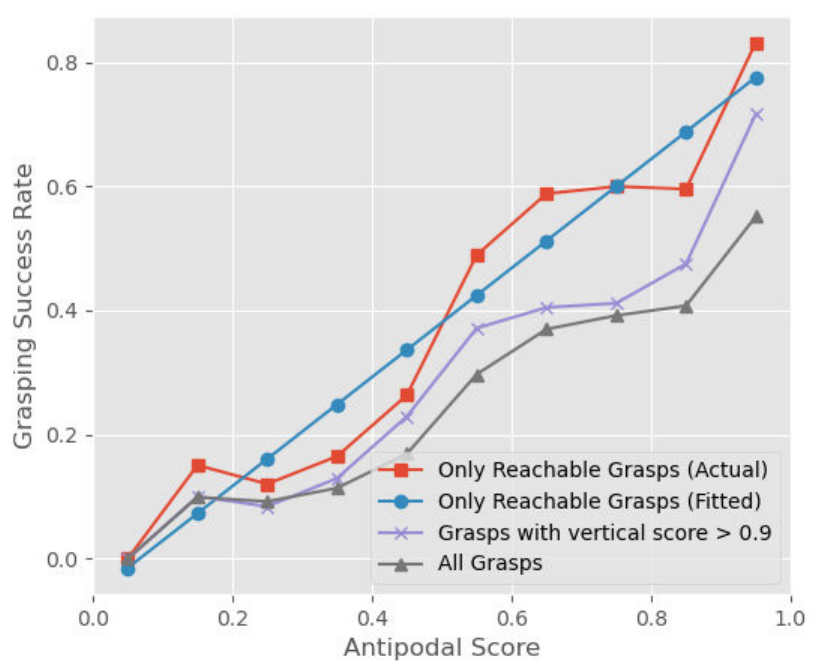}
\label{fig:antipodal_grasp}
}%
\subfigure[]{
\centering
\includegraphics[height=4.5cm,width=5.2cm]{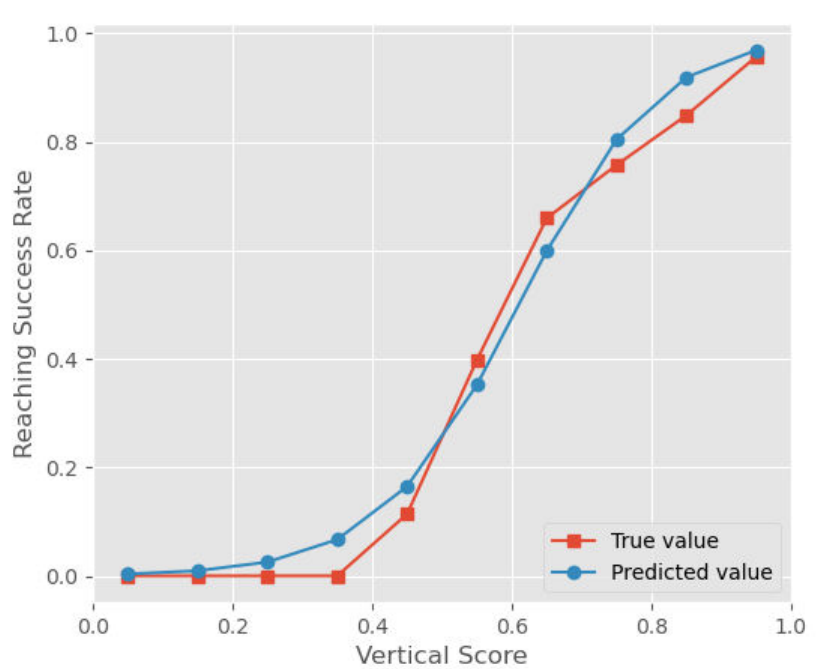}
\label{fig:vertical_reach}
}%
\subfigure[]{
\centering
\includegraphics[height=4.45cm,width=5.7cm]{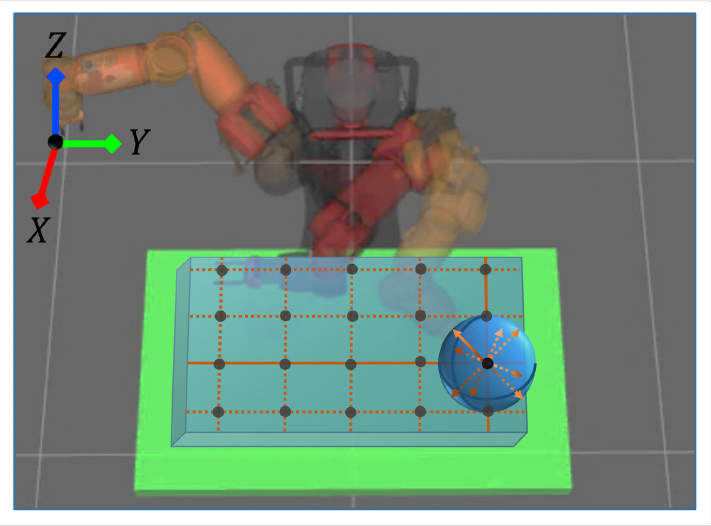}
\label{fig:reach_test}
}%
\caption{(a) Effect of antipodal scores on the grasping success rate.
(b) Effect of vertical scores on the reaching success rate. When the tolerances between the desired and reached positions and poses are both within the set thresholds, we consider this grasp to be successfully reached. (c) The actual robotic experiment setup for the effect of Vertical Score on the Reaching Success Rate.}
\end{figure*}

In RN, we refine grasp proposals close to the ground truth.
Therefore, we formulate this problem into the classification and residue regression problem.
The to-be-predicted residues are calculated as:
$$
res_{Rc} = (c-\hat{c})/c_b
$$
$$
{res}_{Rr} = \frac{\bm{r}}{\|\bm{r}\|} - \frac{\bm{\hat{r}}}{\|\bm{\hat{r}}\|},
$$
$$
res_{Ru} = u-\hat{u}, \ u \in \{\theta,s_q\} 
$$
where $res_{Ru}$ are the ground-truth residues, 
and $u, \ \hat{u}, \ (u \in {c, \bm{r}, \theta, s_q})$ are the ground-truth and predicted center, orientation, approach angle, and quality score in GRN.
We assign a binary class label $y$ (of being close to the ground truth or not) to each proposal.
The positive/negative label is introduced as follows.
$$
y=
   \begin{cases}
   1  &\mbox{$ |c-\hat{c}| < d_1, \ \langle \bm{r}, \bm{\hat{r}} \rangle < \beta_1, \ |\theta-\hat{\theta}| < \gamma_1$}\\
   0  &\mbox{$ |c-\hat{c}| > d_2, \ \langle \bm{r}, \bm{\hat{r}} \rangle \ge \beta_2, \ |\theta-\hat{\theta}| \ge \gamma_2\ $}
   \end{cases}  ,
$$

We minimize the RN loss following the multi-task loss.
We use the cross-entropy loss to classify grasp proposals' labels, and
for proposals close to the ground truth, we directly regress the residues through smooth L1 loss.
The RN loss is formulated as follows.
$$
L_3=\frac{\lambda_{Rcls}}{N_{cls}} \sum_{y_i \in \{ 0, 1 \} } L_{Rcls} (y_i, \hat{y}_i) + \frac{\lambda_{Ru}}{N_{reg}} \sum_{y_i = 1} \sum_u  L_{Ru}
$$
$$
L_{Ru} = SL(res_{Ru}, \hat{res}_{Ru}), \ u \in \{ c, \bm{r}, \theta, s_q\}  \eqno{(4)}, 
$$
where $L_{Rcls}, L_{Ru}$ are the proposal classification loss and the residue regression loss, $y_i, \hat{y}_i$ are the ground-truth and predicted proposal class labels, $res_{Ru}, \hat{res}_{Ru}$ are the ground-truth and predicted residues in RN.
Note that proposals that are neither positive nor negative do not contribute to the training objective.
The term $L_{Ru}$ is activated only for positive proposals ($y_i=1$) and is disabled otherwise.
In practice, we set $\lambda_{Rcls} = 0.2, \lambda_{Ru} = 1$.

\subsection{Optimal Grasp Selection Strategy} \label{Optimal Grasp Selection Strategy}

After \regnet generates a set of grasp proposals, it is essential to design a grasp selection strategy to choose the optimal grasp.
According to the definition of Grasp $g=(c,\bm{r},\theta)\in\mathbb{R}^7$, where $c=(x,y,z)\in \mathbb{R}^3, \bm{r}=(r_x,r_y,r_z)\in \mathbb{R}^3$ and $\theta\in [-{\pi/2},{\pi/2}]$ represent the grasp center, gripper orientation (Y-axis) and approach angle, respectively, it can be inferred that $p(g) = p(g|c,\bm{r},\theta))$. And according to the defination of antipodal score and vertical score as following, we can get that $p(s_q) \propto p(r,c)$ and $p(s_v) \propto p(\theta)$. And then, $p(g) \propto p(s_q, s_v)$. For clarity, let's use $p(grasping)$ instead of $p(g)$.
Different from REGNet which selects the grasps with predicted scores $> threshold$ which is set to 0.5, we propose two strategies based on the predicted grasp quality score (antipodal score) $\hat{s_q}$ and the vertical score $s_v$.
Note that the vertical score $s_v$ can be calculated directly without prediction.

\subsubsection{Heuristic Policy}\ 

Among a bunch of grasp candidates predicted by \regnet, the heuristic policy directly selects the grasp with the largest sum of the predicted antipodal score $\hat{s_q}$ and the vertical score $s_v$ to perform grasping. The following experiments in simulation qualitatively analyze the effects of the antipodal and vertical scores on the grasping success rate.

We use the Franka Emika Panda robot with a parallel-jaw gripper to perform grasping in the Pybullet (\cite{2016_pybullet}) simulation environment.
Specifically, we randomly generate $120$ multi-object scenes and sample more than $1000$ collision-free grasps for each scene.
The Panda robot will attempt each sampled grasp.
The robot cannot reach all positions with a specified pose owing to the limitation of the robot joint angles.

We count different antipodal score intervals and corresponding grasping success rates for all performed grasps (the grey polyline) and reachable grasps (the red polyline), respectively, which is shown in Figure~\ref{fig:antipodal_grasp}.
We calculate that the Pearson correlation coefficient between the antipodal score and grasping success rate is $0.9687$, 
which illustrates that there is a significant positive correlation between them.

In addition, by comparing the grey and red polylines, we find that whether the grasp is reachable also affects the grasping success rate.
Compared to the grasping success rate of all performed grasps, the success rate of grasps with vertical scores higher than $0.9$ (purple polyline) is slightly higher, which demonstrates that a higher vertical score helps improve the grasping success rate. 
Therefore, we directly take the sum of the antipodal and vertical scores as the grasp sampling metric.

\subsubsection{Analytic Policy}\ 

The analytic policy selects the optimal grasp according to the influence degree of the antipodal and vertical score on the grasping success rate.
We offer a quantitative analysis of the respective effects of the antipodal and vertical scores.

To establish the proper mathematical model, we utilize the previous analysis of the effectiveness of $s_q$ and $s_v$ on the grasping success rate and introduce a reaching success rate to simplify the problem. First, we analyze the factors affecting the reaching success rate through the real Baxter robot.
For robots with different physical and dynamic parameters, it is possible that the vertical score $s_v$ has a different effect on the reaching success rate.
Considering the gaps between the simulated and real robots, we directly conduct this experiment based on the Baxter.
We sample a set of 6-DOF poses according to the following rules and let the robot try to reach these poses.
The 6-DOF pose usually consists of a 3d position $(x,y,z)$ and rotation represented by Euler angles $(r_x,r_y,r_z)$.
As shown in Figure~\ref{fig:reach_test}, we sample $100$ different 3D positions by uniformly dividing the operating platform area into small blocks.
And we sample $72$ different rotation angles which cover grasp poses with different vertical scores as much as possible.
Then we record the poses that can be reached and count the reaching success rate of the corresponding vertical score intervals, which is shown in Figure~\ref{fig:vertical_reach}.

When the robot is placed perpendicular to the horizontal (X-Y) plane, the smaller the angle between its end-effector and the Z-axis, the higher the vertical score, leading to a higher reaching success rate, which is shown in Figure~\ref{fig:vertical_reach} and it is similar with Sigmoid function.
Therefore, based on the Sigmoid function fitting, we establish the relationship between the two quantities, which is modeled as $P(reaching) = 1/(1+e^{-10.1244(s_v-0.6103)})$.
Based on the least-squares linear fitting, we model the relationship between the grasping success rate and the antipodal score for those reachable grasps as $P(grasping|reaching) = 0.8783s_q - 0.0587$.
According to the formula of conditional probability, $P(grasping, reaching) = P(grasping|reaching) \cdot P(reaching)$.
Usually, when a grasp executes successfully, it must reach successfully, and hence, we assume that a necessary condition for successfully grasping is successfully reaching, which means that $P(grasping, reaching) = P(grasping)$.
We finally formula the probability of successfully grasping as:
$$
P(grasping) = \frac{0.8783s_q - 0.0587}{1+e^{-10.1244(s_v-0.6103)} }, \eqno{(5)} 
$$
Among a set of predicted grasp candidates, we use the predicted antipodal score $\hat{s_q}$ and vertical score $s_v$ to calculate $P(grasping)$ and select the optimal grasp that maximizes $P(grasping)$.

\subsection{Implementation Details}
In this paper, our network is suitable for generating grasps for different-width grippers.
However, it can be easily extended to situations where the other gripper parameters are different.
The network inputs contain the observed point cloud and the gripper width. 
We randomly sample $N = 25600$ points from the raw point cloud and set the unit of the input gripper width to \textit{m}.

In GRN, we set the number of selected positive points $k_1$ as $768$ when generating grasp regions.
And the grasp region radius $\phi$ is set as $0.02m$ through ablation studies.
We keep $K=256$ points in a grasp region to ensure the fixed-dimensional input.
In practice, $S=0.1, \ M=4$.
When generating ground-truth labels of grasp anchors, we set $\alpha_1 = 2\pi / 3$ and $\alpha_2 = {5\pi}/{12}$.
   
In RN, we keep $64$ points in the gripper closing area.
When generating positive and negative labels, we set $d_1=0.015$, $d_2=0.020$, $\beta_1 = {\pi}/{4}$, $\beta_2 = {\pi}/{3}$, $\gamma_1 = {\pi}/{4}$, and $\gamma_2 = {\pi}/{3}$

All stages are trained simultaneously with a batch size of $6$ and a learning rate of $0.005$ in the beginning. 
The learning rate is divided by $2$ every $5$ epochs.
We utilize the Adam optimizer to solve the optimization problem.
In experiments, our gripper-embedded network is trained for 15 epochs using the multi-gripper dataset, while other gripper-free methods are trained for 30 epochs using the dataset with a gripper width of $0.08$.

\section{Simulation Experiments}
We evaluate \regnet both in simulation and the real world aimed at answering the following questions:
(1) How effective is the gripper-embedded network in detecting grasps for grippers with different sizes relative to several point-cloud-based gripper-free methods?
(2) How well does our method generalize to novel objects and new environments?
(3) How does each component in the network affect the final performance in terms of collision-free rate, antipodal score, and coverage rate?

\begin{table*}[ht]
\small\sf\centering
\caption{Grasp detection performance of different methods tested on Dataset-All in simulation.}
\setlength{\tabcolsep}{2mm}{
\begin{tabular}{lccccccc}
\toprule         
\multirow{2}{*}{\bf Method} & \multicolumn{4}{c}{\bf Grasp Performance} & \multicolumn{3}{c}{\bf Time Efficiency (ms)} \\
\cmidrule(r){2-5} \cmidrule(r){6-8} 
& \bf CFR & \bf AS & \bf AS(w/ C) & \bf TCR    & \bf FPT & \bf PT & \bf TT \\
\midrule
\texttt{Random}        &52.62\textpm0.10\% &0.4134\textpm0.0003 &0.2645\textpm0.0007 &\textbf{23.57\textpm0.02\%}  &\textbf{0.00} &2858.28 &2858.28\\
\texttt{GPD(3 chann)}  &63.57\textpm0.49\% &0.5496\textpm0.0052 &0.4058\textpm0.0065 &6.12\textpm0.49\%    &1.14 &6513.84  &6514.98 \\
\texttt{GPD(12 chann)} &62.33\textpm0.27\% &0.5576\textpm0.0049 &0.4088\textpm0.0043 &3.90\textpm0.61\%    &1.52 &6921.96 &6923.48 \\
\texttt{PointNetGPD}   &65.84\textpm0.23\% &0.5815\textpm0.0042 &0.4388\textpm0.0042 &5.04\textpm0.53\%    &22.14 &2994.90 & 3017.04 \\
\texttt{S4G}           &71.93\textpm0.19\% &0.7662\textpm0.0006 &0.5900\textpm0.0018  &10.46\textpm0.02\%  &168.69 &35.27	&\textbf{203.96}\\
\texttt{REGNet}        &79.84\textpm1.05\% &0.7280\textpm0.0104 &0.6165\textpm0.0098  &6.68\textpm0.35\%   &170.41 &\textbf{33.78} &204.19\\
\texttt{\regnet(layer1)} &\textbf{88.20\textpm0.25\%} &\textbf{0.8174\textpm0.0024} &\textbf{0.7395\textpm0.0029} &8.62\textpm0.07\%  &222.44 &35.05 &257.49\\
\texttt{\regnet(layer2)} &85.68\textpm0.35\% &0.8007\textpm0.0018 &0.7096\textpm0.0033 &19.67\textpm0.77\%  &223.77 &34.93 &258.70\\
\bottomrule
\end{tabular}}
\label{table:total_performace}
\begin{tablenotes}
 \item[1] We optimize the code of collision detection in REGNet (\cite{2021_regnet}) to reduce the \textit{processing time (PT)}.
\end{tablenotes}
\end{table*}

Tested on our simulated dataset, \regnet significantly outperforms several baseline methods for grippers with different sizes.
Likewise, \regnet is tested on REGRAD (\cite{2022_regrad}), a new dataset with novel objects from the ShapeNet dataset (\cite{chang2015shapenet}). For different grippers (even novel ones), it can be successfully generalized to novel objects and environments.
Furthermore, we conduct a series of ablation experiments to ensure the effectiveness of each component.

\subsection{Evaluation Metrics}

The simulation experiments are evaluated on the grasp dataset generated by MuJoCo (\cite{2012_mujoco}) and Blender (\cite{2018_blender}).
Following S4G (\cite{2020_s4g}) and 6-DOF GraspNet (\cite{2019_6dof_graspnet}), we evaluate the grasp detection performance using these metrics:
\begin{enumerate}
\item \textit{collision-free ratio (CFR)}, the proportion of grasps that do not collide with the entire scene.
\item \textit{antipodal score (AS)}, which describe the force closure property of generated grasps;
\item \textit{antipodal score with collision (AS w/ C)}, describing the combined properties of the collision-free ratio and force closure property;
\item \textit{top coverage rate (TCR)}, the diversity of the grasps with top $100$ predicted quality scores;
\end{enumerate}

Note that although both \textit{AS} and \textit{AS(w/ C)} measure the force closure property, \textit{AS} adds antipodal scores of all grasps with or without collision to the sum, while \textit{AS(w/ C)} sets the scores of grasps that are not collision-free as $0$.
\textit{AS(w/ C)} is the weighted sum of \textit{AS} and \textit{CFR}.
\textit{AS} decouples \textit{CFR} from \textit{AS(w/ C)}, and reflects the force closure property more realistically than \textit{AS(w/ C)}.
Figure~\ref{fig:antipodal_grasp} demonstrates the effectiveness of the antipodal score as the quality score $s_q$ for assessing the grasp quality.
The grasping success rate shows a general upward trend with the increase of the antipodal score.

\textit{TCR} measures how well the space of ground-truth grasps is covered by the top $100$ grasps that don't collide with the observed point cloud.
Following 6-DOF GraspNet (\cite{2019_6dof_graspnet}), if there exists a predicted grasp whose center is within $2cm$ from the center of a ground-truth grasp $g \in G$, the grasp $g$ is covered.
The coverage rate is the ratio of the covered grasps to all ground-truth grasps.
To be more specific, for convenience, we perform collision checking with the observed point cloud for $1000$ grasps with top predicted quality scores and select top $100$ grasps with higher scores among them to calculate the above metrics.


Furthermore, we compare the time efficiency of different methods, including the \textit{forward-passing time (FPT)}, \textit{processing time (PT)}, and \textit{total time(TT)}.
\textit{PT} usually contains the pre-processing time for grasp sampling and post-processing time for collision checking with the observed point cloud.
\textit{TT} is the sum of \textit{FPT} and \textit{PT}.

\begin{figure*}
\centering
\includegraphics[height=4cm,width=17.5cm]{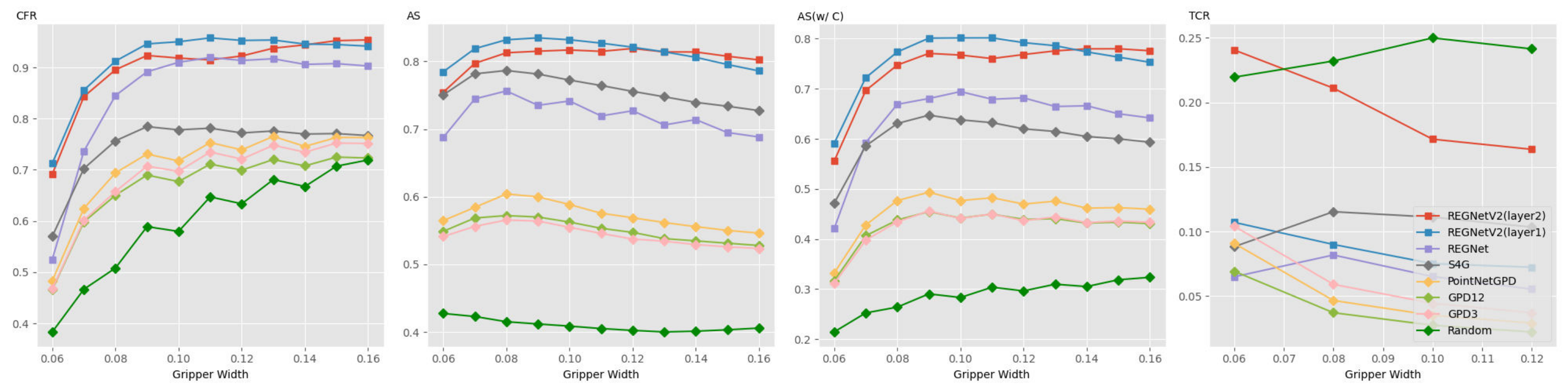}
\caption{Grasp detection performance for grippers with different widths. The X-axis, and Y-axis represent the gripper width and evaluation metrics, respectively. We test the performance on grippers with widths of $0.06m$ to $0.16m$ (with an interval of $0.01m$).}
\label{fig:performace_diffwidth}
\end{figure*}

\subsection{Results and Analysis}

\subsubsection{Baselines}\

The baseline methods include the random sampling method, 3-channel and 12-channel versions of GPD (\cite{2016_gpd}), 2-class single-view PointNetGPD (\cite{2019_pointnetgpd}), S4G (\cite{2020_s4g}), and REGNet (\cite{2021_regnet}). 
For each method, we train $5$ models and measure the mean and variance of the performance.
Except for the methods based on point cloud based scene representation, there are also algorithms that utilize voxel-represented scenes to detect grasp like VGN (\cite{2021vgn}) and EdgeGrasp (\cite{2023edge}). 

In experiments of random sampling method, GPD and PointNetGPD, we randomly sample $1000$ points on the surface of the observed point cloud. 
Taking these points as centers, we generate $100$ grasps that do not collide with the observed point cloud.
In the random sampling method experiment, the entire $100$ grasps are used to evaluate the performance.
In contrast, GPD and PointNetGPD train grasp quality score classifiers and select positive grasps with high scores for performance evaluation. The score threshold is set as $0.6$.
For a fair comparison, in S4G, REGNet, and REGNetV2 algorithms, we perform collision checking with the observed point cloud for top $1000$ grasps with higher predicted quality scores and select top $100$ grasps that do not collide with the observation to evaluate the performance for a fair comparison. 
All baseline methods are trained using the grasp dataset with a width of $0.08m$, while our method uses datasets with widths of $0.06$, $0.08$, $0.10$, and $0.12m$.




\subsubsection{Performance} \

For the above methods, we evaluate the grasp performance on grippers with widths of $0.06m$ to $0.16m$ (with an interval of $0.01m$).
All widths are new parameters except that $0.06$, $0.08$, $0.10$, and $0.12m$ are in the training set.

Table \ref{table:total_performace} illustrates the overall performance of different methods tested on test datasets with widths of $0.06$, $0.08$, $0.10$, and $0.12m$ (Dataset-ALL).
It indicates that our gripper-embedded network, \regnet (layer1 and layer2) is superior to the baselines on different datasets.
However, since the random method randomly generates grasps at any position, while other methods tend to focus on some easy-to-grasp positions to generate grasps, it has a higher \textit{TCR} but lower \textit{CFR} and \textit{AS}.
The last two rows in Table \ref{table:total_performace} show results of our method sampling positive points from the first and second layers, named \regnet(layer1) and \regnet(layer2), respectively.
The points in the second layer are sampled from the first layer based on the farthest point sampling method.
Therefore, keeping the number of sampling points $k_1$ the same, the points sampled from the second layer will cover more diverse locations but are not necessarily the most suitable for grasping.
It leads to a slight decrease in \textit{CFR}, \textit{AS}, and \textit{AS(w/ C)} but a large increase in \textit{TCR} for \regnet(layer2). 

\begin{table}[ht]
\small\sf\centering
\setlength{\tabcolsep}{0.9mm}{
\caption{Grasp detection performance of different methods tested on REGRAD.}
\begin{tabular}{lcccc}
\toprule
\bf method & \bf CFR & \bf AS & \bf AS(w/ C) & \bf TCR\\
\midrule
\texttt{Random}          &24.54\%  &0.0996  &0.0704 &17.69\%  \\
\texttt{GPD(3 chann)}    &25.46\%  &0.1259  &0.0907 &0.32\% \\
\texttt{GPD(12 chann)}   &29.60\%  &0.1396  &0.1014 &0.57\% \\
\texttt{PointNetGPD}     &24.75\%  &0.1155  &0.0898 &0.36\% \\
\texttt{S4G}             &71.31\%  &0.4806  &0.3848 &7.58\% \\
\texttt{REGNet}          &82.21\%  &0.4402  &0.3991 &9.49\% \\
\texttt{\regnet(layer1)}      &\textbf{87.87\%}  &\textbf{0.5076}  &\textbf{0.4802} &12.95\%  \\
\texttt{\regnet(layer2)}      &83.87\%  &0.4572  &0.4237 &\textbf{27.55\%} \\
\bottomrule
\end{tabular}
\label{table:performace_regrad}
}
\end{table}

We also compare the time efficiency of different methods.
It is not easy to estimate normal accurately from the observed point cloud, causing the low efficiency of sampling grasps based on the Darboux frame (\cite{2020_s4g}). 
Therefore, our end-to-end algorithm is more efficient than the algorithms that sample first and then evaluate, such as Random, GPD, and PointNetGPD.
Compared to REGNet and S4G, the forward-passing time of \regnet increases slightly due to its more complex network structure.
Our method is able to process about five scenes per second with a 3090 GPU.

Moreover, Figure~\ref{fig:performace_diffwidth} demonstrates that \regnet significantly outperforms all baselines for grippers with different widths, even though the width is novel.
It can detect grasps for different grippers through \regnet.
Especially for narrower grippers, \regnet performs better and increases the \textit{CFR} and \textit{AS} by more than $14\%$ and $0.03$ relative to S4G.
As for evaluating \textit{TCR}, we only count it for the gripper whose width is in the generated dataset ($0.06$, $0.08$, $0.10$, and $0.12m$).
Consistent with Table \ref{table:total_performace}, \regnet(layer1) has higher \textit{CFR}, \textit{AS} and \textit{AS(w/ C)} in most cases, while \regnet(layer2) has higher \textit{TCR}.
In addition, we can also see that grippers with different widths will perform differently in the same scene.
In general, the grasp performance of the gripper with moderate width is higher.

Table \ref{table:vgn baseline} shows the results compared with voxel-based algorithm VGN. We evaluate the model under their metrics where it selects random one of the grasps with score $> \epsilon$ until two consecutive failures and then computes the \textit{Grasp Success Rate} and \textit{Declutter Rate}.
 Table \ref{table:vgn baseline} illustrates the model trained by REGNet dataset $G$ has a higher grasp success rate, but the declutter rate is poor.

\begin{table}[ht]
\small\sf\centering
\setlength{\tabcolsep}{0.9mm}{
\caption{Grasp detection performance in simulator used in \cite{2021vgn}}
\begin{tabular}{lccc}
\toprule
\bf method & \bf pile($5$objs) & \bf packed($5$objs) & \bf pile($10$objs)\\
\midrule
\texttt{VGN($\epsilon_1$)}   &65.4/41.6  &89.5/85.9 &59.4/25.1  \\
\texttt{VGN($\epsilon_2$)}   &62.3/46.4  &87.6/90.1 &59.3/34.6  \\
\texttt{VGN($\epsilon_3$)}   &59.8/51.1  &85.8/89.5 &52.8/30.1  \\
\texttt{our dataset($\epsilon_1$)}   &98.1/21.1 &98.1/26.4 &94.2/11.5 \\
\texttt{our dataset($\epsilon_2$)}   &94.5/22.7 &95.3/27.4 &92.2/13.9 \\
\texttt{our dataset($\epsilon_3$)}   &93.2/21.7 &94.3/27.1 &92.1/12.4  \\
\bottomrule
\end{tabular}
\label{table:vgn baseline}
\begin{tablenotes}
 \item[1]$\epsilon_1=0.95$, $\epsilon_2=0.90$, $\epsilon_2=0.80$.
\end{tablenotes}
}
\end{table}


\subsubsection{Generalization to Novel Objects and Environments}\

We evaluate the performance of \regnet on a new dataset REGRAD (\cite{2022_regrad}), which has novel object models and a different distribution relative to our simulated dataset. In addition, we train and evaluate our network on a standard object dataset GraspNet 1 Billion and dataset used in EdgeGrasp Network.
We randomly select $1000$ observed point clouds from each dataset for the test.
Table \ref{table:performace_regrad} and Table \ref{table:different dataset} show that our method is also effective for grasp detection on novel objects and different environments in 3D space.

\begin{table}[ht]
\small\sf\centering
\setlength{\tabcolsep}{0.9mm}{
\caption{Grasp detection performance on different datasets.}
\begin{tabular}{lcccc}
\toprule
\bf method & \bf CFR & \bf AS & \bf AS(w/ C) & \bf TCR\\
\midrule
\texttt{REGNet dataset}          &88.20\%  &\textbf{0.8174}  &\textbf{0.7395} &8.62\% \\
\texttt{REGRAD dataset}          &87.87\%  &0.5076  &0.4802 &12.95\% \\
\texttt{EdgeGrasp dataset}          &\textbf{99.80\%}  &0.3035  &0.2916 &\textbf{52.58\%} \\
\texttt{GraspNet 1 Billion}      &93.67\%  &0.3786  &0.3575 &31.69\%  \\
\bottomrule
\end{tabular}
\label{table:different dataset}
}
\end{table}

\begin{figure}
\centering
\includegraphics[height=7.8cm,width=8.3cm]{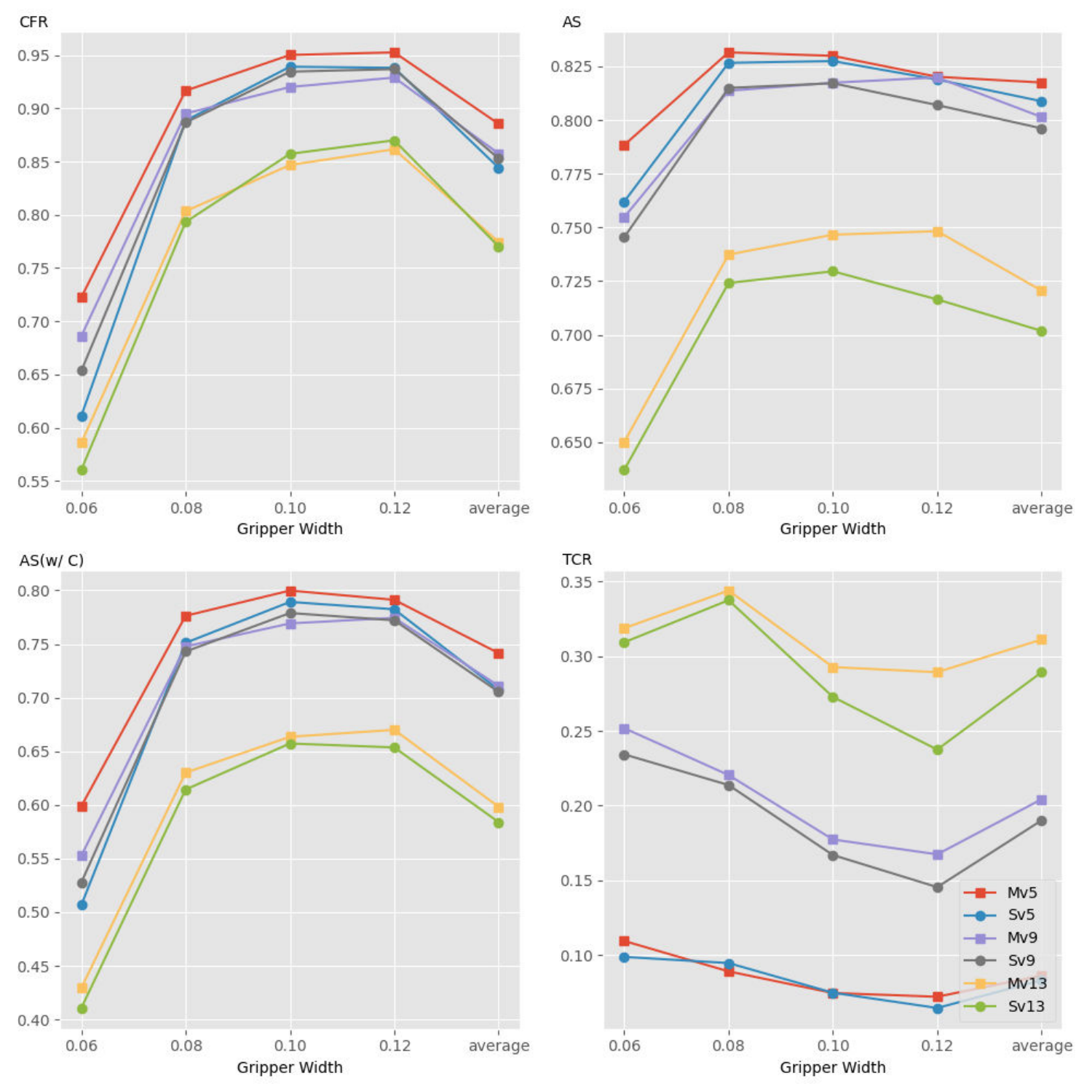}
\caption{The performance of REGNet and \regnet.
M/Sv5, M/Sv9, and M/Sv13 are versions of sampling positive points from the first, second, and third layers, respectively.
The rightmost column calculates the average performance on test datasets with different widths (datasets-All).}
\label{fig:gripper_contrast}
\end{figure}

\subsubsection{Ablation Studies} \ 

We designed the following experiments to evaluate the effectiveness of several main components in our network.

\textbf{Effectiveness of the Gripper-embedded Input.}
As mentioned before, \regnet, the gripper-embedded network embeds the parameters of the gripper and observed point cloud to the inputs for grasp prediction. 
In this way, it can generate collision-free grasps with high grasp quality scores for grippers of different sizes.
In contrast, REGNet is a gripper-free network with most of the structure consistent with \regnet but only trained on the dataset with the width of $0.08m$.

As shown in Figure~\ref{fig:gripper_contrast}, we compare the performance of REGNet and \regnet on datasets with different gripper widths.
The results show that \regnet can achieve higher performance than REGNet no matter which layer the positive points are sampled from.
The performance of REGNet drops significantly when generalized to narrower grippers, such as those with a width of $0.06m$.
However, in \regnet, the gripper-embedded architecture helps learn the gripper size and generate grasps for different grippers.
Moreover, the number of parameters (Params.) and floating-point operations per second (FLOPs) of \regnet are only slightly higher than REGNet, which is shown in Table \ref{table:ablation}.

\begin{table*}[ht]
\small\sf\centering
\caption{Ablation analysis results tested on Dataset-ALL.}
\setlength{\tabcolsep}{2.2mm}{
\begin{tabular}{lccccccccccc}
\toprule
version \\ REGNet & layer & rates & anchor & region & refine & \#Params. & \#FLOPs & CFR & AS & AS w/ CFR & TCR  \\
\midrule
\texttt{Sv1}     &1  &100.\% &\xmark  &\xmark &\xmark  &\textbf{5.545M}  &168.252G  &81.06\%  &0.7690 &0.6644  &10.66\% \\
\texttt{REGNet}  &1  &0.25\% &\cmark  &\cmark  &\cmark  &6.514M  &157.537G &81.06\% &0.7409  &0.6335  &6.44\%  \\
\midrule
\texttt{Sv2}    &1  &3.00\%  &\xmark  &\xmark  &\xmark  &5.907M  &150.728G  &76.43\%  &0.7497 &0.6158  &\textbf{11.64\%}  \\
\texttt{Sv3}    &1  &3.00\%  &\cmark  &\xmark  &\xmark  &6.274M  &150.735G  &78.55\%  &0.7708 &0.6425  &8.20\%  \\
\texttt{Sv4}    &1  &3.00\%  &\cmark  &\cmark  &\xmark  &6.274M  &150.735G  &80.37\%  &0.7857 &0.6639  &8.19\%  \\
\texttt{Sv5}    &1  &3.00\%  &\cmark  &\cmark  &\cmark  &6.514M  &244.713G  &\textbf{84.42\%}  &\textbf{0.8088} &\textbf{0.7073}  &8.31\%  \\
\midrule
\texttt{Sv6}    &2  &15.0\%  &\xmark  &\xmark  &\xmark  &5.907M  &150.728G  &79.62\%  &0.7247 &0.6179 &\textbf{28.50\%}   \\
\texttt{Sv7}    &2  &15.0\%  &\cmark  &\xmark  &\xmark  &6.274M  &150.735G  &80.60\%  &0.7659 &0.6553  &18.83\%  \\
\texttt{Sv8}    &2  &15.0\%  &\cmark  &\cmark  &\xmark  &6.274M  &150.735G  &80.40\%  &0.7698 &0.6566 &18.84\%  \\
\texttt{Sv9}    &2  &15.0\%  &\cmark  &\cmark  &\cmark  &6.514M  &244.713G  &\textbf{85.28\%}  &\textbf{0.7961} &\textbf{0.7054}  &19.01\%  \\
\midrule
\texttt{Sv10}   &3  &75.0\%  &\xmark  &\xmark  &\xmark  &5.907M  &150.728G  &71.07\%  &0.5891 &0.4699  &\textbf{31.41\%}   \\
\texttt{Sv11}   &3  &75.0\%  &\cmark  &\xmark  &\xmark  &6.274M  &150.735G  &67.86\%  &0.6297 &0.4856  &23.49\%  \\
\texttt{Sv12}   &3  &75.0\%  &\cmark  &\cmark  &\xmark  &6.274M  &150.735G  &72.90\% &0.6715 &0.5407 &27.26\%   \\
\texttt{Sv13}   &3  &75.0\%  &\cmark  &\cmark  &\cmark  &6.514M  &244.713G  &\textbf{77.02\%}  &\textbf{0.7018}  &\textbf{0.5840} &28.93\%  \\
\bottomrule
\end{tabular}\\[10pt]}

\setlength{\tabcolsep}{2.0mm}{
\begin{tabular}{lccccccccccc}
\toprule
version \\ \regnet \quad & layer & rates & anchor & region & refine & \#Params. & \#FLOPs & CFR & AS & AS w/ CFR & TCR \\
\midrule
\texttt{Mv1}    &1  &100.\%  &\xmark  &\xmark  &\xmark  &5.546M  &168.542G  &66.83\%  &0.6383 &0.5510  &8.20\%  \\
\midrule
\texttt{Mv2}    &1  &3.00\%  &\xmark  &\xmark  &\xmark  &5.909M  &151.019G  &83.06\%  &0.7653 &0.6640 &\textbf{12.71\%} \\
\texttt{Mv3}    &1  &3.00\%  &\cmark  &\xmark  &\xmark  &6.276M  &151.026G  &85.60\%  &0.7982 &0.7090  &8.27\%  \\
\texttt{Mv4}    &1  &3.00\%  &\cmark  &\cmark  &\xmark  &6.276M  &151.026G  &86.05\%  &0.8019 &0.7130  &8.42\%  \\
\texttt{Mv5}    &1  &3.00\%  &\cmark  &\cmark  &\cmark  &6.515M  &245.004G  &\textbf{88.54}\%  &\textbf{0.8175} &\textbf{0.7415} &8.63\% \\
\midrule
\texttt{Mv6}    &2  &15.0\%  &\xmark  &\xmark  &\xmark   &5.909M  &151.019G  &81.01\%  &0.7314 &0.6291 &\textbf{29.59\%}  \\
\texttt{Mv7}    &2  &15.0\%   &\cmark  &\xmark  &\xmark  &6.276M  &151.026G  &81.73\%  &0.7777 &0.6679 &20.31\%  \\
\texttt{Mv8}    &2  &15.0\%   &\cmark  &\cmark  &\xmark  &6.276M  &151.026G  &83.34\%  &0.7854 &0.6845 &19.80\%  \\
\texttt{Mv9}    &2  &15.0\%   &\cmark  &\cmark  &\cmark  &6.515M  &245.004G  &\textbf{85.75\%}  &\textbf{0.8014} &\textbf{0.7110} &20.43\%  \\
\midrule
\texttt{Mv10}   &3  &75.0\%  &\xmark  &\xmark  &\xmark  &5.909M  &151.019G  &72.90\%  &0.6076 &0.4932 &\textbf{34.64\%}  \\
\texttt{Mv11}   &3  &75.0\%  &\cmark  &\xmark  &\xmark  &6.276M  &151.026G  &74.86\%  &0.6778 &0.5546  &31.88\%  \\
\texttt{Mv12}   &3  &75.0\%  &\cmark  &\cmark  &\xmark  &6.276M  &151.026G  &74.52\%  &0.6910 &0.5607  &29.88\%   \\
\texttt{Mv13}   &3  &75.0\%  &\cmark  &\cmark  &\cmark  &6.515M  &245.004G  &\textbf{77.44\%}  &\textbf{0.7206} &\textbf{0.5982}  &31.12\% \\
\bottomrule
\end{tabular}
\label{table:ablation}
}
\end{table*}

\textbf{Effectiveness of Positive Point Selecting Mechanism.}
In GRN, increasing the number of selected positive points $k_1$ will increase the calculation time but can obtain higher \textit{TCR}.
Keeping $k_1$ the same will guarantee a consistent computation time for a fair comparison.
Since the PointNet++ encoder uses FPS to sample points, sampling positive points in deeper layers can cover a more diverse space.
If we would like to get a higher \textit{TCR}, we need to sample positive points at a deeper layer while keeping $k_1$ consistent.

We design a series of experiments to demonstrate the effectiveness of the positive point-selecting mechanism.
As illustrated in Table \ref{table:ablation}, S/Mv2-5, S/Mv6-9, S/Mv10-13 are the versions of REGNet and \regnet, which sample $k_1$ positive points from the first, second, third layers, respectively, with sampling rates of $3\%$, $15\%$ and $75\%$.
Since the number of points in different layers is different, when $k_1$ is set as $768$, different layers have different sampling rates.

The top coverage rate \textit{TCR} of versions sampling points from deeper layers are much higher than sampling from the first layer (S/Mv5), while S/Mv5 have a higher \textit{CFR}, \textit{AS} and \textit{AS(w/ C)}.
In addition, we find that the performance of versions that sampling points from the third layer (S/Mv13) is much lower.
That is because the third layer covers the observation with fewer points, resulting in not all selected positive points being suitable for grasping, and the regression result centered on these points will be poor.

\textbf{Effectiveness of the Grasp Anchor Mechanism.}
We design the w/o anchor and w/ anchor versions, whose difference is whether to use the anchor mechanism or not.
These versions are all designed with different sampling layers.
S/Mv3, S/Mv7, and S/Mv11 utilize the grasp anchor mechanism, while S/Mv2, S/Mv6, and S/Mv10 directly regress proposals without classifying the category of grasp anchors.
In Table \ref{table:ablation}, most versions with the anchor have better grasp detection performance than those without the anchor, which demonstrates that the grasp anchor mechanism contributes inextricably to the performance.

\textbf{Effectiveness of the Grasp Region.}
For a fair comparison, we design the w/o region and w/ region versions, 
which regress grasps from the single point and grasp region features, respectively.
Compared with S/Mv3, S/Mv7, and S/Mv11, keeping the sampling layer the same, S/Mv4, S/Mv8, and S/Mv12 regress proposals using the grasp region features instead of single-point features.
As illustrated in Table \ref{table:ablation}, the performance of most versions degrades without the grasp region, which indicates the effectiveness of the grasp region.
Moreover, we compare the performance of different radii of grasp regions in \regnet to select a proper sphere radius. 
The results are shown in Table \ref{table:result_radius}. 
After the comprehensive comparison, we set the radius $\phi$ of the grasp region as $0.02m$.

\begin{table}[ht]
\setlength{\tabcolsep}{1.65mm}{
\small\sf\centering
\caption{Results of setting different grasp region radius tested on Dataset-ALL.}
\begin{tabular}{lccccc}
\toprule
\bf versions & \bf radius & \bf CFR & \bf AS & \bf AS(w/ C) & \bf TCR \\
\midrule
\texttt{Mv3} &\texttt{w/o}   &85.60\%  &0.7982 &0.7090 &8.27\% \\
\texttt{Mv4} &\texttt{0.01m} &85.78\%  &0.7982 &0.7095 &\textbf{8.80\%}  \\
\texttt{Mv4} &\texttt{0.02m} &\textbf{86.05\%}  &\textbf{0.8019} &\textbf{0.7130} &8.42\% \\
\texttt{Mv4} &\texttt{0.03m} &82.50\%  &0.7790 &0.6731 &8.63\% \\

\midrule
\texttt{Mv7} & \texttt{w/o}  &81.73\%  &0.7777 &0.6679  &\textbf{20.31\%}\\
\texttt{Mv8} & \texttt{0.01m} &82.32\%  &\textbf{0.7867} &0.6781 &20.11\% \\
\texttt{Mv8} & \texttt{0.02m} &\textbf{83.34\%} &0.7854  &\textbf{0.6845} &19.80\% \\
\texttt{Mv8} & \texttt{0.03m} &81.07\%  &0.7756 &0.6623  &19.57\% \\

\bottomrule
\end{tabular}
\label{table:result_radius}
}
\end{table}

\begin{figure}
\centering
\includegraphics[height=5.4cm,width=8.4cm]{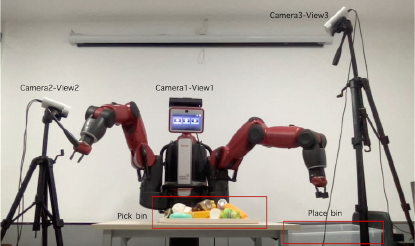}
\caption{Environment Setup.
}
\label{fig:env}
\end{figure}

\textbf{Effectiveness of the Refine Network.}
To analyze the efficiency of the Refine Network, we compare the w/o RN and w/ RN versions, whose difference is containing RN or not. 
Table \ref{table:ablation} shows that RN facilitates improvements in grasp detection performance.
The versions containing the Refine Network (S/Mv5, S/Mv9, S/Mv13) all achieve the highest results on the indicator \textit{CFR} and \textit{AS},
but their \textit{TCR} may decrease slightly because negative proposals after classification are discarded in RN stage.

\textbf{Effectiveness of Using RGB Features.}
To analyze the influence of RGB features on grasp detection performance, we train and test \regnet using point clouds with and without RGB information.
Except for the number of the input channel, versions (w/ RGB) have the same architecture as versions (w/o RGB).
As shown in Table \ref{table:performace_rgb}, versions using RGB features generally have a higher performance.

\section{Real Robotic Experiments}
To assess the grasp performance and generalization ability of several algorithms under novel objects, stacked clutters, and different camera models and viewpoints in the real world,
Our real robotic experiments include several diverse test scenes, which are (1) grasping in single-object scenes, (2) bin-picking, and (3) grasping via varied camera perspectives.
Grasping in single-object scenes is a simple test to obtain the grasp detection performance for isolated objects in the real world.
Bin-picking experiments evaluate the performance of grasping objects in stacked scenes.
We devise the grasping experiments under varied camera perspectives to evaluate the adaptability of our algorithm relative to camera perspectives.
All experiments are tested on Baxter with 2-finger parallel-jaw grippers, which is shown in Figure~\ref{fig:env}.

\begin{table}[ht]
\small\sf\centering
\setlength{\tabcolsep}{1.65mm}{
\caption{The influence of RGB features on the grasp detection performance of \regnet.}
\begin{tabular}{lcccc}
\toprule
\bf versions & \bf CFR & \bf AS & \bf AS (w/ C) & \bf TCR\\
\midrule
\texttt{Mv5 (w/\quad\,RGB)}   &\textbf{88.54\%}  &\textbf{0.8175}  &\textbf{0.7415} &8.63\% \\
\texttt{Mv5 (w/o RGB)}        &88.30\%  &0.8092  &0.7347 &\textbf{8.90\%} \\
\midrule
\texttt{Mv9 (w/\quad\,RGB)}   &\textbf{85.75\%}  &\textbf{0.8014}  &\textbf{0.7110} &20.43\% \\
\texttt{Mv9 (w/o RGB)}        &85.22\%  &0.7857  &0.6956 &\textbf{21.00\%} \\
\bottomrule
\end{tabular}
\label{table:performace_rgb}
}
\end{table}

\begin{figure}[thbp]
\centering
\subfigure[Food items.]{
\centering
\includegraphics[height=3.1cm,width=3.8cm]{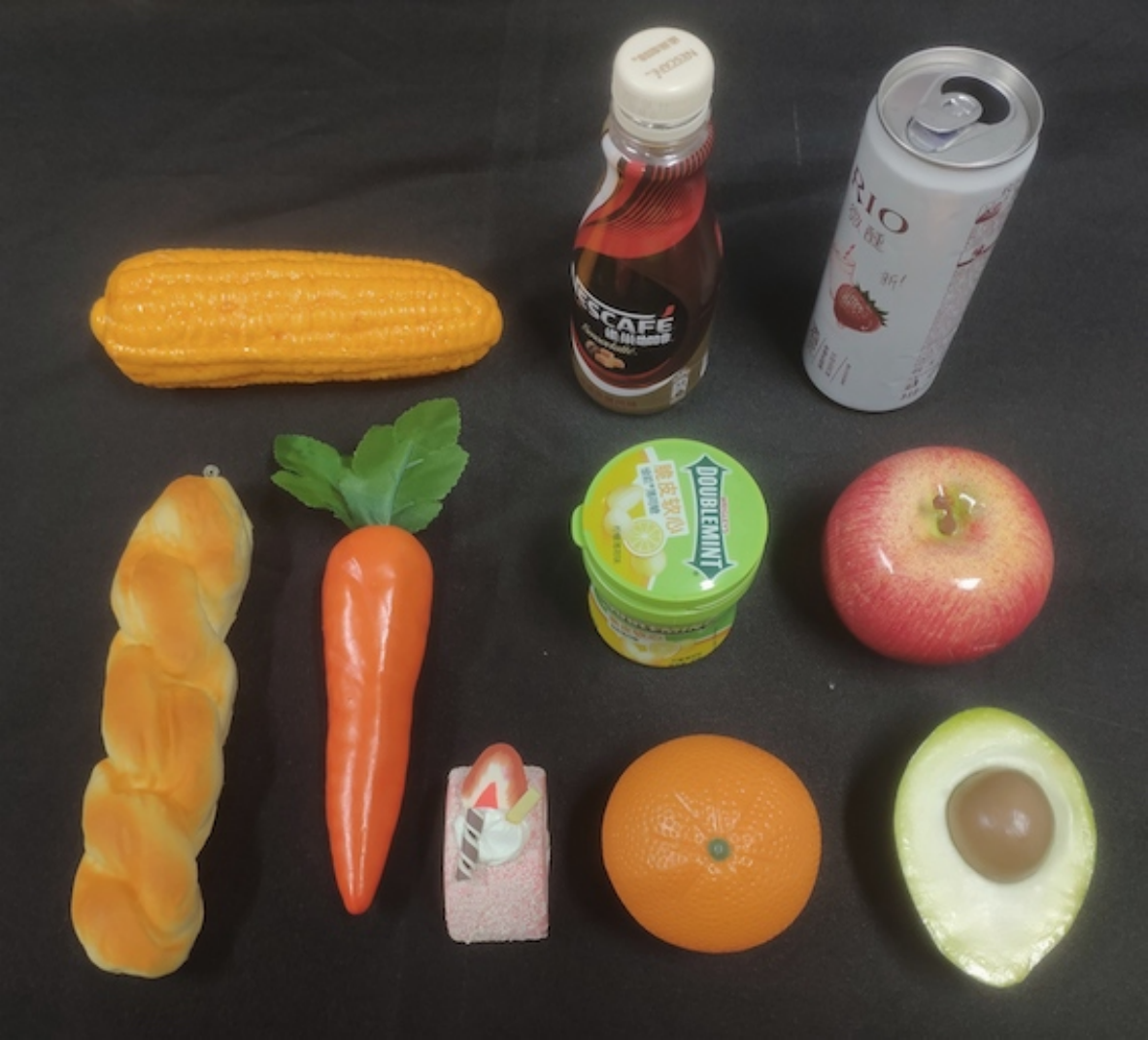}
\label{fig:Food}
}%
\subfigure[Toy items.]{
\centering
\includegraphics[height=3.1cm,width=3.8cm]{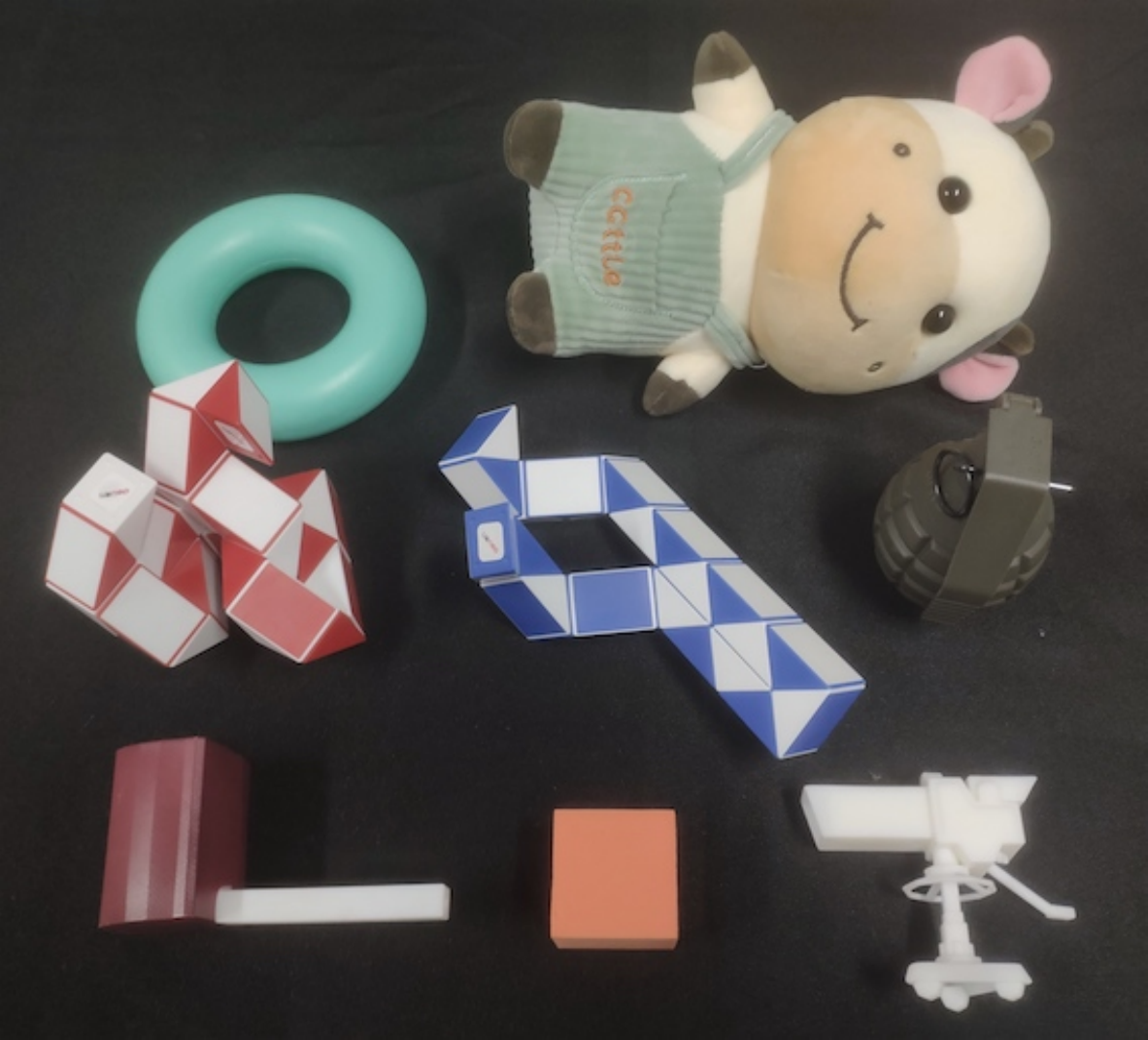}
\label{fig:Toy}
}%

\subfigure[Tool items.]{
\centering
\includegraphics[height=3.1cm,width=3.8cm]{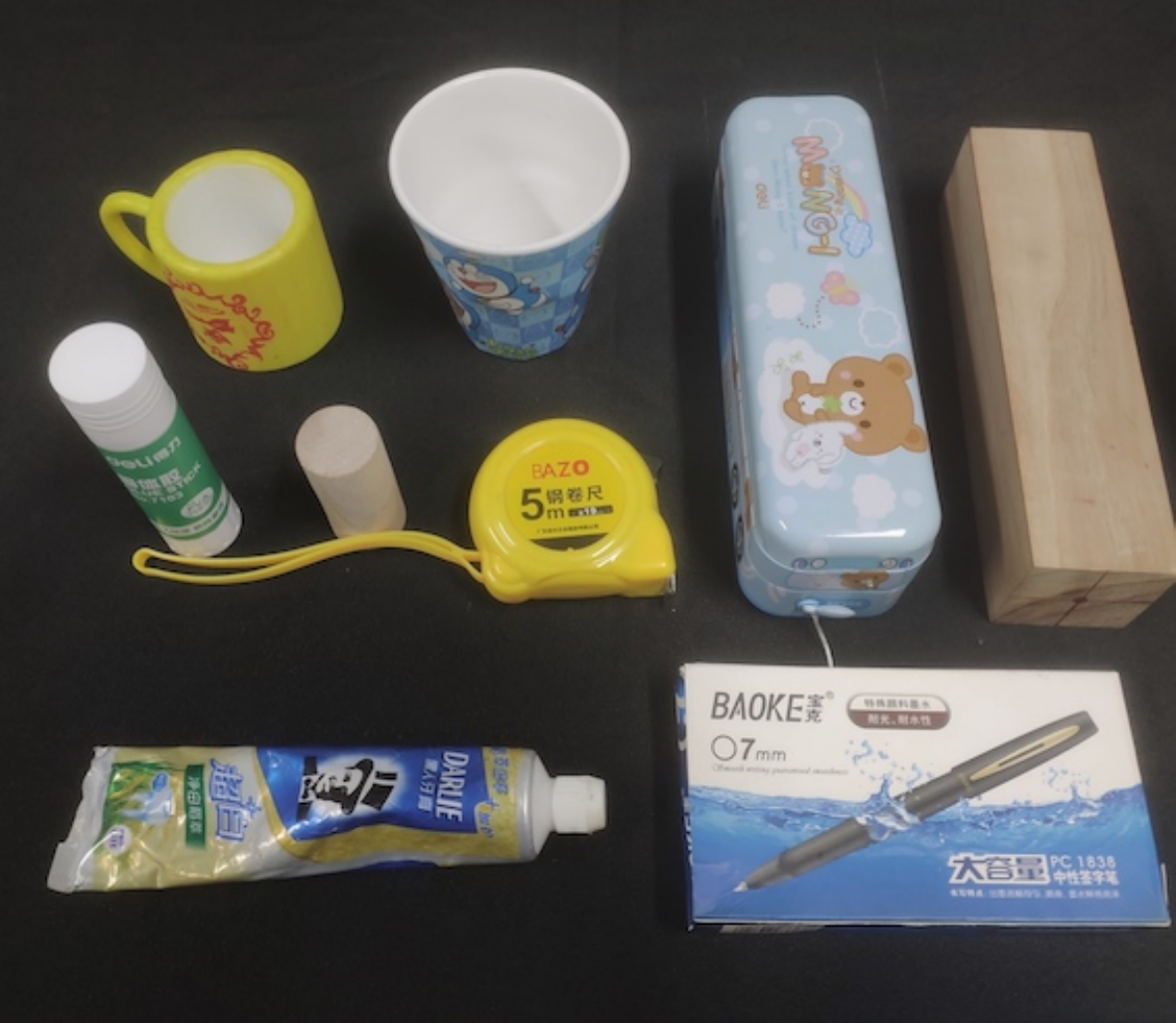}
\label{fig:Tool}
}%
\subfigure[Shape items.]{
\centering
\includegraphics[height=3.1cm,width=3.8cm]{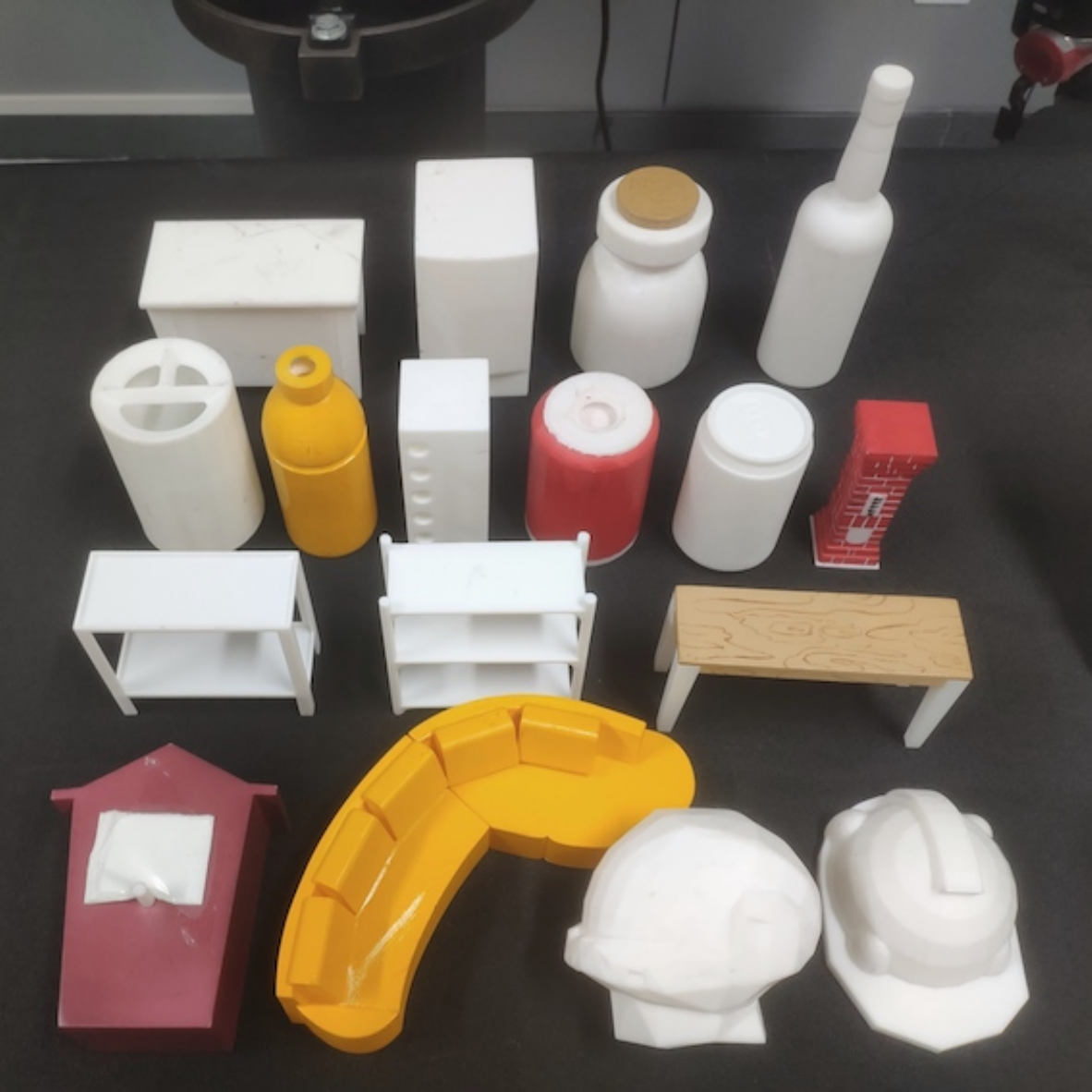}
\label{fig:Shape}
}%
\caption{The set of objects used in the real robotic experiments.}
\label{fig:dataset_real}
\end{figure}

\subsection{Evaluation Metrics}
Following the previous literature, we adopt the \emph{grasping success rate (SR)} and \emph{completion rate (CR)} as grasping performance evaluation metrics.

The success rate is the ratio of successful grasps among all performed grasps. 
A grasp is considered to be a success if the manipulator grasps the object and places it in the place bin (Figure~\ref{fig:env}).
For the case of grasping multiple objects at the same time, we treat it as a failure.
For multi-object scenes, the completion rate is the percentage of objects successfully removed from the pick bin relative to the total number of objects used to evaluate completeness.

\subsection{Grasping In Single-object Scenes}
This section provides the experiments of grasping isolated objects in the real world.
For grippers with different widths, our method outperforms all baselines and is able to generalize to novel objects.

\subsubsection{Experiment Setup} \

In the experiments of grasping in single-object scenes, the point clouds used for detection are observed by the KinectV2 mounted on the head of Baxter. 

We select novel objects not in our training dataset for real robotic experiments.
Our algorithm is trained based on the YCB object set, which contains $5$ different categories.
Referring to the YCB set, we employ $44$ objects divided into $4$ categories of food, toy, tool, and shape items.
They are visualized in Figure~\ref{fig:dataset_real}.

\begin{figure*}
\centering
\includegraphics[height=6cm,width=17.6cm]{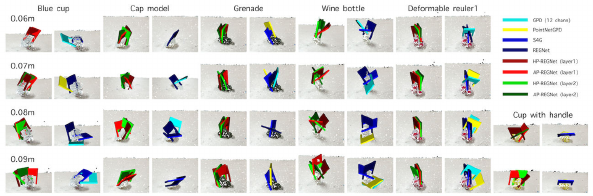}
\caption{Comparison between to-be-executed grasps predicted by different methods in the experiments of grasping in single-object scenes.
Every two columns are under the same object and observation.
A sub-figure without grasps represents that the current method does not detect collision-free grasps in the current observation.
}
\label{fig:single_grasp}
\end{figure*}

\textbf{Food items.} They contain common food models and product boxes, basically in the shape of cylinders and spheres.
From left to right and top to bottom, the objects are corn, coffee bottle, Rio can, bread, carrot, gum jar, apple, small cake, orange, and avocado, respectively.

\begin{table}[ht]
\small\sf\centering
\setlength{\tabcolsep}{1.5mm}{
\caption{Success rate of single-object grasping experiments.}
\begin{tabular}{lcccc}
\toprule
\bf methods                 &\bf 0.06m &\bf 0.07m&\bf 0.08m &\bf 0.09m\\
\midrule
\texttt{GPD (12 chann)}     &7/20  &8/20   &7/20  &9/20 \\
\texttt{PointNetGPD}        &8/20  &9/20   &9/20  &8/20 \\
\texttt{S4G}                &13/20  &13/20   &12/20  &13/20 \\
\texttt{REGNet}             &15/20  &14/20   &13/20  &14/20 \\
\texttt{HP-\regnet(layer1)} &\textbf{17/20}  &16/20   &\textbf{16/20}  &\textbf{17/20} \\
\texttt{AP-\regnet(layer1)} &\textbf{17/20}  &17/20   &\textbf{16/20}  &\textbf{17/20} \\
\texttt{HP-\regnet(layer2)} &17/20  &16/20   &14/20  &16/20 \\
\texttt{AP-\regnet(layer2)} &16/20  &\textbf{18/20}   &14/20  &15/20 \\
\bottomrule
\end{tabular}
\begin{tablenotes}
 \item[1] 1. The performance of GPD3 and GPD12 is similar, so we don't repeat the experiments for GPD3.
 \item[2] 2. HP-\regnet and AP-\regnet use the same\regnet model but different grasp selection strategies, heuristic, and analytic policy, respectively.
\end{tablenotes}
\label{table:performace_single}
}
\end{table}

\textbf{Toy items.} They include some toys in daily life.
In order from left to right and top to bottom, they are plastic ring, plush toy, deformable ruler1, deformable ruler2, grenade, mailbox, sponge block, and gun model.
Some are complex and non-convex in shapes, such as deformable rulers, plastic grenades, and plastic guns.

\textbf{Tool items.} In order from left to right and top to bottom, this set contains a cup with a handle, a blue cup, a stationery case, woodblock, glue stick, cylindrical woodblock, band tape, toothpaste, and pen refill packaging.
The blue plastic cup is characterized by a smooth surface, which is not easy to grasp.

\textbf{Shape items.} They consist of $17$ 3D-printed objects described in the REGRAD (\cite{2022_regrad}) dataset. 
From left to right and top to bottom, they are block1, block2, medicine bottle, wine bottle, cylinder1, yellow bottle, block3, red can, cylinder2, block4, shelf1, shelf2, shelf3, house model, sofa model, helmet, and cap model. 
The back two rows in Figure~\ref{fig:Shape} contain objects with irregular and curved geometric shapes, while the front rows are mostly regular cuboids and cylinders.

\begin{table}[ht]
\small\sf\centering
\setlength{\tabcolsep}{3.1mm}{
\caption{Grasping success rate of different object items. }
\begin{tabular}{lcccc}
\toprule
\bf Items       &\bf Obj Num &\bf First 4 &\bf Last 4 &\bf Total \\
\midrule
\texttt{Food}   &10 &63.60\%   &92.51\%  &77.69\% \\
\texttt{Toy}    &8  &26.42\%   &61.27\%  &45.30\% \\
\texttt{Tool}   &9  &43.39\%   &68.24\%  &56.02\% \\
\texttt{Shape}  &17 &60.11\%   &85.81\%  &72.63\% \\
\bottomrule
\end{tabular}
\begin{tablenotes}
 \item[1] The last three columns are the average success rate of the first four baseline algorithms, the last four algorithms, and all algorithms, respectively.
\end{tablenotes}
\label{table:item_average_performance}
}
\end{table}

The maximum width of Baxter's gripper can be set manually by adjusting the screw position, which is suitable for evaluating grasp detection performance for grippers with different geometries.
We adjust the gripper width to $0.06$, $0.07$, $0.08$, and $0.09m$ for the single-object grasping experiment.
For grippers with different widths, we choose a subset of objects from Figure~\ref{fig:dataset_real} that fit the current width. 
An experiment for one algorithm consists of $20$ grasp trials.

\subsubsection{Results and Analysis}\ 

As demonstrated in Table \ref{table:performace_single}, \regnet (layer1) achieves the highest grasping success rate for grippers with widths of $0.06$, $0.08$ and $0.09m$.
And \regnet (layer2) using analytic grasp selection strategy achieves the highest success rate for the gripper with a width of $0.07m$.
It suggests that \regnet can effectively generate stable grasps despite being trained only on synthetic data. 
However, compared with the heuristic grasp selection strategy, the analytic policy has no clear advantage in the impact of the success rate.

We count the average grasping success rate of the $4$ object items.
As illustrated in Table \ref{table:item_average_performance}, in general, the toy and tool items are more challenging to grasp, mainly due to the complex shape and slippery surface.
Compared with the baseline methods, \regnet (layers 1 and 2) significantly improve the grasp detection performance for objects with complex or non-convex shapes.

Figure~\ref{fig:single_grasp} shows some grasp detection results under grippers with different widths using different algorithms in real-world single-object scenarios. 
As mentioned in the previous literature, sampling-based methods, such as GPD and PointNetGPD, sample grasp candidates based on the Darboux frames of the observed surface points. 
Nevertheless, it is challenging to estimate the Darboux frame from noisy point clouds accurately, such as the observation of cups and deformable rulers, resulting in low efficiency.
\begin{figure}[thbp]
\subfigure[Failure case of unstable grasps.]{
\includegraphics[height=3.5cm,width=7.7cm]{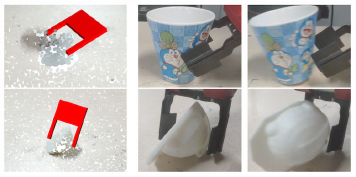}
\label{fig:fail_single1}
}%

\subfigure[Failure case of collision.]{
\includegraphics[height=1.7cm,width=7.7cm]{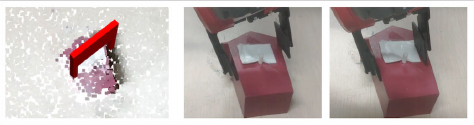}
\label{fig:fail_single2}
}%

\subfigure[Failure case due to the gripper's limitation.]{
\includegraphics[height=1.7cm,width=7.7cm]{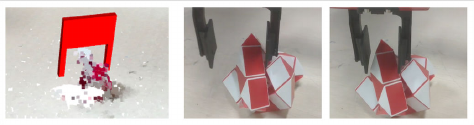}
\label{fig:fail_single3}
}

\caption{Failure cases in experiments of grasping in single-object scenes.}
\label{fig:fail_single}
\end{figure}
Moreover, sampling-based methods are incredibly dependent on the sampled points, which results in that they can only generate grasps on one side of the object with a low coverage rate when the observation is incomplete.
When the positions of the robot and the camera relative to the object are not on the same side, the motion planning of the grasp detected by the sampling-based method often fails.
In contrast, our method can generate properer and relatively complete grasps from noisy and incomplete observation.

\subsubsection{Failure Cases}\ 

We summarize the reasons for grasping failures in single-object grasping experiments as follows. Table \ref{table:fail_single} demonstrates the probability of each reason in failure cases.

\textbf{(1) Unstable grasps.}
This condition is usually caused by the unstable grasps generated by grasp detection algorithms.
The grasp instability may be due to a low antipodal score of the predicted grasp.
Likewise, when grasping an object with slippery materials or curved surfaces, even a grasp with a high antipodal score may not be force-closure, resulting in the grasping failure.
In Figure~\ref{fig:fail_single1}, grasps detected for the blue plastic cup and the cap model both have high antipodal scores.
However, when grasping the slippery cup and cap, they accidentally slip out of the gripper (in the last column).

Table \ref{table:fail_single} indicates that $65.14\%$ of the failure cases are due to this reason.
The analytic grasp selection strategy effectively helps to select the top-quality grasps.

\textbf{(2) Colliding while grasping.}
Though we perform the collision checking for those detected grasps with the observation, it is still possible for them to collide with the entire scene due to the incomplete point cloud observed.
When the size of the object to be grasped is close to the gripper, minor errors in the camera's intrinsic and extrinsic matrices may cause the collision.

\begin{table}[ht]
\small\sf\centering
\setlength{\tabcolsep}{0.5mm}{
\caption{The number of grasping failure cases and the proportion of failure reasons. }
\begin{tabular}{lcccc}
\toprule
\bf methods & \bf Num &\bf Instability &\bf Collision &\bf Gripper \\
\midrule
\texttt{HP-\regnet(layer1)}  &14 &71.43\%  &7.14\%   &21.43\%  \\
\texttt{AP-\regnet(layer1)}  &13 &53.85\%  &23.08\%  &23.08\%  \\
\texttt{HP-\regnet(layer2)}  &17 &76.47\%  &17.65\%  &5.88\%  \\
\texttt{AP-\regnet(layer2)}  &17 &58.82\%  &23.53\%  &17.65\%  \\
\texttt{Average}             &15 &65.14\%  &17.85\%  &17.01\%  \\
\bottomrule
\end{tabular}
\label{table:fail_single}
}
\end{table}

\begin{table}[ht]
\small\sf\centering
\setlength{\tabcolsep}{3.3mm}{
\caption{The average grasping success rate of the selected objects for bin-picking experiments in different items. }
\begin{tabular}{lcccc}
\toprule
\bf Items       &\bf All objs &\bf Easy-10 &\bf Hard-10 &\bf 20 objs\\
\midrule
\texttt{Food}   &77.69\% &82.47\% &63.64\% &78.70\% \\
\texttt{Toy}    &45.30\% &/       &40.63\% &40.63\%\\
\texttt{Tool}   &56.02\% &100.0\% &38.10\% &69.04\%\\
\texttt{Shape}  &72.63\% &79.86\% &44.03\% &63.93\%\\
\bottomrule
\end{tabular}
\begin{tablenotes}
 \item[1] There are no suitable objects in toy items that can be selected as easy-to-grasp objects.
\end{tablenotes}
\label{table:select_obj_difficulty}
}
\end{table}

\textbf{(3) The limitation of the robotic end-effector.}
The Baxter's end-effector cannot wholly be close, and hence, it restricts the minimum size of the object part to be grasped.
Figure~\ref{fig:fail_single3} shows the failure case caused by the limitation of Baxter's gripper.

\subsection{Bin-Picking}\ 
We produce bin-picking experiments to verify our method's effectiveness and generalization ability for detecting grasps in complex real-world clutters.
Compared with other methods, \regnet achieves the highest grasping success rate and completion rate in three designed complex scenarios with different objects and distributions.
In stacked scenes with 20 objects, AP-\regnet (layer1) can achieve a success rate of $74.98\%$ and a completion rate of $87.00\%$.

\begin{table*}[ht]
\small\sf\centering
\caption{The comparison of grasping performance of bin-picking experiments in various scenarios.}
\setlength{\tabcolsep}{4.5mm}{
\begin{tabular}{lcccccc}
\toprule         
\multirow{2}{*}{\bf Method} & \multicolumn{2}{c}{\bf Easy - 10 objects} & \multicolumn{2}{c}{\bf Hard - 10 objects} & \multicolumn{2}{c}{\bf 20 objects} \\
\cmidrule(r){2-3} \cmidrule(r){4-5}  \cmidrule(r){6-7} 
& \bf SR & \bf CR & \bf SR & \bf CR & \bf SR & \bf CR \\
\midrule
\texttt{GPD(12 chann)}      &49.97\% &72.00\% &46.39\% &58.00\% &32.00\%* &40.00\%*\\
\texttt{PointNetGPD}        &51.43\% &76.00\% &48.00\% &58.00\% &36.00\%* &45.00\%*\\
\texttt{S4G}                &83.64\% &78.00\% &61.69\% &76.00\% &66.97\% &73.00\%\\
\texttt{REGNet}            &86.32\% &84.00\% &63.97\% &82.00\% &70.06\% &79.00\% \\
\texttt{HP-\regnet(layer1)} &85.39\% &92.00\% &65.71\% &78.00\% &74.47\% &82.00\%\\
\texttt{AP-\regnet(layer1)} &\textbf{91.21\%} &94.00\% &\textbf{69.75\%} &80.00\% &\textbf{74.98\%} &87.00\%\\
\texttt{HP-\regnet(layer2)} &86.21\% &\textbf{98.00\%} &66.88\% &\textbf{86.00\%} &73.47\% &\textbf{95.00\%}\\
\texttt{AP-\regnet(layer2)} &85.91\% &96.00\% &66.87\% &84.00\% &73.92\% &90.00\%\\
\bottomrule
\end{tabular}}
\label{table:performace_bin_picking}
\begin{tablenotes}
 \item[1] * Owing to the higher grasping failure probability using GPD (12 channels) and PointNetGPD methods in complex stacked clutters, we only perform one trial in 20-object scenes.
\end{tablenotes}
\end{table*}

\begin{figure*}
\centering
\includegraphics[height=10.1cm,width=17.6cm]{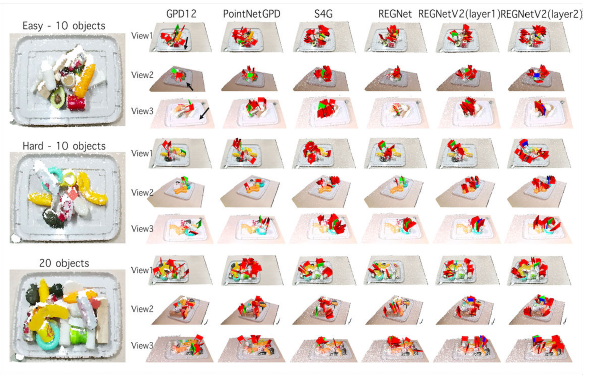}
\caption{
The visualization of $30$ randomly chosen predicted grasps under three different camera perspectives in various scenes.
The red grasps are part of predicted grasps, while the green and blue grasps are the to-be-executed grasps selected by heuristic and analytic policy, respectively.
The black arrows in the first $3$ rows indicate the raycasting direction under different camera perspectives.
}
\label{fig:multi_grasp}
\end{figure*}

\subsubsection{Experiment Setup}\ 

In the experiments of bin-picking, we still utilize the KinectV2 mounted on the head of Baxter to observe the point clouds (View 1).
We utilize the gripper with a width of $0.07m$ to execute grasping.

We design $3$ scene configurations with different objects and densities to perform experiments.
They are scenes with (1) $10$ easy-to-grasp objects (Easy-$10$ objects), (2) $10$ hard-to-grasp objects (Hard-$10$ objects), (3) $20$ mixed objects ($20$ objects), respectively.
Each method is tested for $5$ times in each configuration.
For one trial in the scene containing $N$ objects, we provide $N+5$ grasping attempts.
The Baxter executes grasping until no grasp is detected or $N+5$ opportunities are exhausted.

Regardless of the grasping detection method used, since the grasping success rate is related to object types to a certain extent, it is essential to find an efficient and fair manner to establish the hierarchy of objects' grasping difficulty and then select suitable easy-to-grasp and hard-to-grasp objects.
Different from the method of sampling empirical $75^{th}$ percentile of grasp quality described in  \cite{2020_egad} and \cite{2019_adversarial_grasp}, we estimate the grasping difficulty by calculating the average grasping success rate tested in single-object grasping experiments.
There is a negative correlation between the grasping difficulty and the success rate.
Then we select $10$ easy-to-grasp and hard-to-grasp objects that fit the gripper with a width of $0.07m$ respectively.
The grasping success rate of the selected three object sets is illustrated in Table \ref{table:select_obj_difficulty}.
The average success rate of the $20$ mixed objects has a similar distribution compared to the formal object items.

\subsubsection{Results and Analysis}\ 

\textbf{Effectiveness of \regnet}.
Table \ref{table:performace_bin_picking} demonstrates the grasping performance of different methods under the $3$ types of scene configurations.
AP-\regnet (layer1) achieves the highest grasping success rate, which is $91.21\%$ in the 10-easy-object scene, $69.75\%$ in the 10-hard-object scene and $74.98\%$ in the 20-object scene.
And HP-\regnet (layer2) achieves the highest completion rate in these scenarios.
The performance of GPD12 and PointNetGPD decreases significantly compared to the end-to-end methods.
As shown in Figure~\ref{fig:multi_grasp}, the approach directions of grasps detected by GPD12 and PointNetGPD are disorganized due to the defective estimated Darboux frames.

The \regnet version of sampling positive points from the second layer covers a more diverse space for grasp detection.
Figure~\ref{fig:multi_grasp} shows that in most cases, \regnet (layer2) guarantees a higher grasp coverage rate, which is able to generate grasp for different objects.
After many grasping trials, only a few small-sized objects remain in the scene.
Under the circumstances, the captured point cloud is difficult to describe the geometric information of objects due to the camera error.
Therefore, the grasp confidence of the observed points predicted from the ScoreNet of \regnet is small. 
\regnet (layer1) tends to focus on the sampling positive points in a local region, which may lead to the failure of grasp detection, thus ending the experiment.
Nevertheless, the sampled positive points of \regnet (layer2) are more dispersed, which gives a chance to detect suitable grasps of small-sized objects, leading to a higher completion rate in the bin-picking experiment.
In addition, \regnet (layer1 and 2) both sample $768$ positive points to regress grasp proposals, while REGNet only samples $64$ points, and hence, the completion rate drops partially.

Although the 20-object scene contains all objects in the 10-easy-object and 10-hard-object scenes, its grasping performance is lower than the average performance in the 10-object scene.
After failing to grasp an object successfully, a stable algorithm usually generates the optimal grasps for the same object several times.
Therefore, the grasping performance of hard-to-grasp objects makes a more significant impact on multi-object scenarios. 
We need to improve the lower bound of grasp detection performance for hard-to-grasp objects.

\begin{figure}
\centering
\includegraphics[height=10.5cm,width=8.5cm]{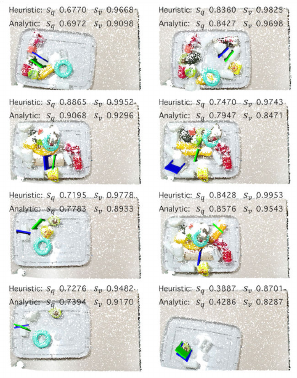}
\caption{
The visualization of to-be-executed grasps selected by heuristic (green) and analytic policy (blue).
The caption of each sub-figure describes the predicted antipodal score and the vertical score of the selected grasp.
}
\label{fig:policy}
\end{figure}

\textbf{Effectiveness of the analytic grasp selection mechanism}.
Table \ref{table:performace_bin_picking} demonstrates the effectiveness of the analytic grasp selection mechanism. 
Relative to the HP-\regnet (layer1), the success rate of AP-\regnet (layer1) increases by $5.82\%$, $4.04\%$, and $0.51\%$ respectively under the three scene configurations.
And most of the grasping performance results using the analytic strategy are superior to the heuristic strategy.
The grasps selected by the heuristic and analytic policy are sometimes almost identical.
We display some different selected grasps in Figure~\ref{fig:policy}.
The predicted antipodal score $s_q$ of the grasp selected by the analytic policy is often higher than the heuristic policy in the same scenario.

\subsubsection{Failure Cases}\ 

\begin{figure}[thbp]
\subfigure[Failure case of imprecise grasps.]{
\includegraphics[height=7.4cm,width=8cm]{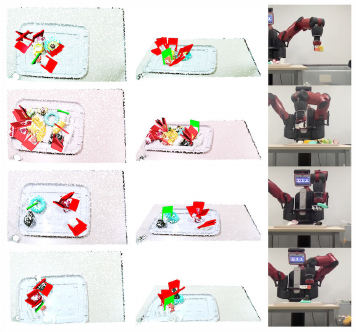}
\label{fig:fail_multi1}
}%

\subfigure[Failure case due to the camera's limitation.]{
\includegraphics[height=5cm,width=8cm]{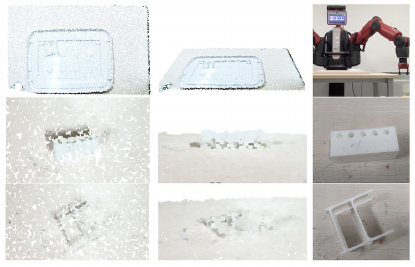}
\label{fig:fail_multi2}
}%

\caption{Failure cases in bin-picking experiments. The green grasps are the selected grasps to be executed. }
\label{fig:fail_multi}
\end{figure}
The failure cases in the bin-picking experiment include the reasons mentioned in the single-object scene experiments, akin to the generated grasps are not stable enough, collisions occurring during the motion planning, and the robotic gripper is limited.
Furthermore, imprecise detection of grasp and the limitation of the camera may lead to the failure of grasping, which is shown in Figure~\ref{fig:fail_multi}.

\textbf{(1) Imprecise grasps.}
As described in the section on evaluation metrics, grasping multiple objects simultaneously is treated as a failure.
The imprecise grasp detection is the primary cause for grasping multiple objects as an "atomic" object, which is visualized in the last two rows of Figure~\ref{fig:fail_multi1}.
More specifically, our algorithm does not perform object segmentation, which may detect a single grasp for multiple objects.
Furthermore, in the first two rows of Figure~\ref{fig:fail_multi1}, \regnet successfully detects the grasps.
However, owing to the close distance between multiple objects and the tiny error of the extrinsic camera matrix, the robot inevitably grasps two objects simultaneously.
Through previous experiments, we find that such imprecise grasping usually occurs in detecting small or flat objects.

\begin{table*}[ht]
\small\sf\centering
\caption{The comparison of grasping performance in stacked clutters containing $20$ objects under different camera views.}
\setlength{\tabcolsep}{3.25mm}{
\begin{tabular}{lcccccccc}
\toprule         
\multirow{2}{*}{\bf Method} & \multicolumn{2}{c}{\bf View1} & \multicolumn{2}{c}{\bf View2} & \multicolumn{2}{c}{\bf View3} & \multicolumn{2}{c}{\bf Average}\\
\cmidrule(r){2-3} \cmidrule(r){4-5}  \cmidrule(r){6-7} \cmidrule(r){8-9} 
& \bf SR & \bf CR & \bf SR & \bf CR & \bf SR & \bf CR & \bf SR & \bf CR \\
\midrule
\texttt{HP-\regnet(layer1)} &74.47\% &82.00\% &70.13\% &82.00\% &\textbf{74.32\%} &82.00\% &72.98\% &82.00\%\\
\texttt{AP-\regnet(layer1)} &\textbf{74.98\%} &87.00\% &\textbf{73.88\%} &\textbf{84.00\%} &74.01\% &81.00\% &\textbf{74.29\%} &84.00\%\\
\texttt{HP-\regnet(layer2)} &73.47\% &\textbf{95.00\%} &70.57\% &83.00\% &71.62\% &83.00\% &71.89\% &\textbf{87.00\%}\\
\texttt{AP-\regnet(layer2)} &73.92\% &90.00\% &69.49\% &82.00\% &71.43\% &\textbf{87.00\%} &71.61\% &86.33\%\\
\bottomrule
\end{tabular}}
\label{table:performace_3views}
\end{table*}

\textbf{(2) The limitation of the camera.}
The tabletop observed by the camera is theoretically flat, but in reality, it is a slightly undulating surface, usually due to camera lens distortion, electronic component limitations, etc.
And for some tiny objects with little thickness or depth, it is challenging to predict grasps due to their geometry and indistinct boundary depth, which is illustrated in Figure~\ref{fig:fail_multi2}.

\subsection{Grasping Via Various Camera Perspectives}
Applying robotic grasping algorithms in real-world environments inevitably requires them to adapt to varied camera perspectives.
Compared to RGB images, point clouds are reliable in revealing objects' size and stereoscopic geometric information in diverse observation modalities.
Can our algorithm exploit the observed geometric shapes with size descriptions to detect grasps via different viewpoints?
To answer this question, we pick out other two 6-Dof poses different from the KinectV2 in previous experiments to place cameras.

\subsubsection{Experiment Setup} \

As shown in Figure~\ref{fig:env}, we capture the point clouds using the Kinect DK placed on the side of the table. 
The height difference between the workbench and the camera in the simulated environment is about $0.9m$ during grasp dataset generation.
In real-world experiments, the height differences between the workbench and the three cameras are $0.84$, $0.47$, and $0.93m$, respectively.
These cameras cover as many view angles as possible, among which the second camera with a view angle of nearly $45^{\circ}$ is prone to observe the occlusion.
The experiments are set up in the stacked scenes with $20$ objects, with a similar methodology to the bin-picking experiments.

\subsubsection{Results and Analysis} \

The results in Table \ref{table:performace_3views} illustrate that AP-\regnet (layer1) achieves a grasping success rate of $74.98\%$, $73.88\%$ and $74.01\%$ under the three camera perspectives, respectively.
The success rate of HP-\regnet (layer1) under the third viewpoint is $74.32\%$, which is slightly higher than AP-\regnet (layer1).
Compared with the first and third viewpoints, the second viewpoint's grasp performance is slightly degraded due to the occlusion among objects.
It reveals that \regnet can adapt to several different camera models and perspectives.
Furthermore, AP-\regnet (layer1) has the highest average success rate under the several camera perspectives, reaching $74.29\%$, demonstrating the effectiveness of \regnet with the analytic policy in grasp detection.

\section{Conclusion}
We propose \regnet, an end-to-end region-based grasp network to grasp diversified objects in real-world unstructured environments, which can adapt to different grippers.
To support different grippers, \regnet embeds the gripper parameters into the input point clouds, based on which it predicts proper grasp configurations.
\regnet embeds the gripper parameters into the input point clouds to support different grippers, based on which it predicts proper grasp configurations.
It consists of three stages: Score Network, Grasp Region Network, and Refine Network, which significantly improve the grasp detection performance in dense clutter.
Among the predicted grasp proposals, we devise an analytic policy to select the optimal grasp to be executed.
Furthermore, \regnet is trained on synthetic simulated data with different grippers and can successfully detect grasp configurations on novel objects for diversified grippers in real-world scenarios.


\begin{acks}
This work was supported in part by National Key R\&D Program of China under grant No. 2021ZD0112700,NSFC under grant No.62125305, No.62088102, and No.61973246,the Fundamental Research Funds for the Central Universities under Grant xtr072022001.
\end{acks}

\bibliographystyle{SageH}
\bibliography{ijrr}

\begin{thebibliography}{73}
\providecommand{\natexlab}[1]{#1}
\providecommand{\url}[1]{\texttt{#1}}
\providecommand{\urlprefix}{URL }
\expandafter\ifx\csname urlstyle\endcsname\relax
  \providecommand{\doi}[1]{DOI:\discretionary{}{}{}#1}\else
  \providecommand{\doi}{DOI:\discretionary{}{}{}\begingroup
  \urlstyle{rm}\Url}\fi

\bibitem[{Ainetter and Fraundorfer(2021)}]{2021_end_trainable}
Ainetter S and Fraundorfer F (2021) End-to-end trainable deep neural network
  for robotic grasp detection and semantic segmentation from rgb.
\newblock In: \emph{IEEE International Conference on Robotics and Automation
  (ICRA)}. IEEE, pp. 13452--13458.

\bibitem[{Asif et~al.(2018)Asif, Tang and Harrer}]{2018_graspnet_low_eff}
Asif U, Tang J and Harrer S (2018) Graspnet: An efficient convolutional neural
  network for real-time grasp detection for low-powered devices.
\newblock In: \emph{IJCAI}. pp. 4875--4882.

\bibitem[{Asif et~al.(2019)Asif, Tang and Harrer}]{2019_DSGD}
Asif U, Tang J and Harrer S (2019) Densely supervised grasp detector (dsgd).
\newblock In: \emph{Proceedings of the AAAI Conference on Artificial
  Intelligence}, AAAI'19/IAAI'19/EAAI'19. AAAI Press.
\newblock ISBN 978-1-57735-809-1.
\newblock \doi{10.1609/aaai.v33i01.33018085}.
\newblock \urlprefix\url{https://doi.org/10.1609/aaai.v33i01.33018085}.

\bibitem[{Bohg et~al.(2014)Bohg, Morales, Asfour and
  Kragic}]{data_driven_survey}
Bohg J, Morales A, Asfour T and Kragic D (2014) Data-driven grasp synthesis --
  a survey.
\newblock In: \emph{IEEE Transactions on Robotics}, volume~30. pp. 289--309.

\bibitem[{Borst et~al.(2004)Borst, Fischer and Hirzinger}]{gws1}
Borst C, Fischer M and Hirzinger G (2004) Grasp planning: How to choose a
  suitable task wrench space.
\newblock In: \emph{IEEE International Conference on Robotics and Automation
  (ICRA)}.

\bibitem[{Breyer et~al.(2021{\natexlab{a}})Breyer, Chung, Ott, Siegwart and
  Nieto}]{VGN}
Breyer M, Chung JJ, Ott L, Siegwart R and Nieto J (2021{\natexlab{a}})
  Volumetric grasping network: Real-time 6 dof grasp detection in clutter.
\newblock In: \emph{Conference on Robot Learning}. PMLR, pp. 1602--1611.

\bibitem[{Breyer et~al.(2021{\natexlab{b}})Breyer, Chung, Ott, Siegwart and
  Nieto}]{2021vgn}
Breyer M, Chung JJ, Ott L, Siegwart R and Nieto J (2021{\natexlab{b}})
  Volumetric grasping network: Real-time 6 dof grasp detection in clutter.
\newblock In: \emph{Conference on Robot Learning}. PMLR, pp. 1602--1611.

\bibitem[{Cai et~al.(2022)Cai, Cen, Wang and Wang}]{VPN}
Cai J, Cen J, Wang H and Wang MY (2022) Real-time collision-free grasp pose
  detection with geometry-aware refinement using high-resolution volume.
\newblock \emph{IEEE Robotics and Automation Letters} 7(2): 1888--1895.

\bibitem[{Calli et~al.(2017)Calli, Singh, Bruce, Walsman, Konolige, Srinivasa,
  Abbeel and Dollar}]{2017_ycb}
Calli B, Singh A, Bruce J, Walsman A, Konolige K, Srinivasa S, Abbeel P and
  Dollar AM (2017) Yale-cmu-berkeley dataset for robotic manipulation research.
\newblock In: \emph{The International Journal of Robotics Research (IJRR)},
  volume~36. pp. 261--268.

\bibitem[{Chang et~al.(2015)Chang, Funkhouser, Guibas, Hanrahan, Huang, Li,
  Savarese, Savva, Song, Su et~al.}]{chang2015shapenet}
Chang AX, Funkhouser T, Guibas L, Hanrahan P, Huang Q, Li Z, Savarese S, Savva
  M, Song S, Su H et~al. (2015) Shapenet: An information-rich 3d model
  repository.
\newblock \emph{arXiv preprint arXiv:1512.03012} .

\bibitem[{Chu et~al.(2018)Chu, Xu and Vela}]{2018_multigrasp}
Chu F, Xu R and Vela PA (2018) Real-world multiobject, multigrasp detection
  3(4): 3355--3362.

\bibitem[{Community(2018)}]{2018_blender}
Community BO (2018) \emph{Blender - a 3D modelling and rendering package}.
\newblock Blender Foundation, Stichting Blender Foundation, Amsterdam.
\newblock \urlprefix\url{http://www.blender.org}.

\bibitem[{Coumans and Bai(2016)}]{2016_pybullet}
Coumans E and Bai Y (2016) \emph{Pybullet, a python module for physics
  simulation, games, robotics and machine learning.}
\newblock \urlprefix\url{http://pybullet.org/}.

\bibitem[{Depierre et~al.(2018)Depierre, Dellandrea and
  Chen}]{2018_jacquard_dataset}
Depierre A, Dellandrea E and Chen L (2018) Jacquard: A large scale dataset for
  robotic grasp detection.
\newblock In: \emph{IEEE International Conference on Intelligent Robots and
  Systems (IROS)}. IEEE, pp. 3511--3516.

\bibitem[{Fang et~al.(2020)Fang, Wang, Gou and Lu}]{2020_graspnet_1billion}
Fang H, Wang C, Gou M and Lu C (2020) Graspnet-1billion: A large-scale
  benchmark for general object grasping.
\newblock In: \emph{Proceedings of the IEEE/CVF Conference on Computer Vision
  and Pattern Recognition}. pp. 11444--11453.

\bibitem[{Ferrari and Canny(1992)}]{1992_planning_optimal_grasp}
Ferrari C and Canny J (1992) Planning optimal grasps.
\newblock In: \emph{IEEE International Conference on Robotics and Automation
  (ICRA)}. p. 2290–2295.

\bibitem[{Gschwandtner et~al.(2011)Gschwandtner, KwittAndreas and
  Pree}]{2011_blensor}
Gschwandtner M, KwittAndreas R and Pree U (2011) Blensor: Blender sensor
  simulation toolbox.
\newblock In: \emph{International Symposium on Visual Computing (ISVC)}. pp.
  199--208.

\bibitem[{Gualtieri et~al.(2016)Gualtieri, Ten~Pas, Saenko and
  Platt}]{2016_gpd}
Gualtieri M, Ten~Pas A, Saenko K and Platt R (2016) High precision grasp pose
  detection in dense clutter.
\newblock In: \emph{IEEE International Conference on Intelligent Robots and
  Systems (IROS)}. IEEE, pp. 598--605.

\bibitem[{Hager et~al.(2021)Hager, Bauer, Toussaint and
  Mainprice}]{2021_graspme}
Hager J, Bauer R, Toussaint M and Mainprice J (2021) Graspme-grasp manifold
  estimator.
\newblock In: \emph{IEEE International Conference on Robot \& Human Interactive
  Communication (RO-MAN)}. IEEE, pp. 626--632.

\bibitem[{Huang et~al.(2023{\natexlab{a}})Huang, Wang, Zhu, Walters and
  Platt}]{edgegrasp}
Huang H, Wang D, Zhu X, Walters R and Platt R (2023{\natexlab{a}}) Edge grasp
  network: A graph-based se (3)-invariant approach to grasp detection.
\newblock In: \emph{2023 IEEE International Conference on Robotics and
  Automation (ICRA)}. IEEE, pp. 3882--3888.

\bibitem[{Huang et~al.(2023{\natexlab{b}})Huang, Wang, Zhu, Walters and
  Platt}]{2023edge}
Huang H, Wang D, Zhu X, Walters R and Platt R (2023{\natexlab{b}}) Edge grasp
  network: A graph-based se (3)-invariant approach to grasp detection.
\newblock In: \emph{2023 IEEE International Conference on Robotics and
  Automation (ICRA)}. IEEE, pp. 3882--3888.

\bibitem[{Jiang et~al.(2011)Jiang, Moseson and Saxena}]{2011_rectangle}
Jiang Y, Moseson S and Saxena A (2011) Efficient grasping from rgbd images:
  Learning using a new rectangle representation.
\newblock In: \emph{IEEE International Conference on Robotics and Automation
  (ICRA)}. p. 3304–3311.

\bibitem[{Kamon et~al.(1996)Kamon, Flash and Edelman}]{1996_visual_grasp}
Kamon I, Flash T and Edelman S (1996) Learning to grasp using visual
  information.
\newblock In: \emph{IEEE International Conference on Robotics and Automation
  (ICRA)}. p. 2470–2476.

\bibitem[{Kirkpatrick et~al.(1992)Kirkpatrick, Mishra and Yap}]{gws2}
Kirkpatrick D, Mishra B and Yap CK (1992) Quantitative steinitz's theorems with
  applications to multifingered grasping.
\newblock In: \emph{Discrete \& Computational Geometry}, volume~7. pp.
  295--318.

\bibitem[{Kumra and Kanan(2017)}]{2016_dcnn}
Kumra S and Kanan C (2017) Robotic grasp detection using deep convolutional
  neural networks.
\newblock In: \emph{2017 IEEE/RSJ International Conference on Intelligent
  Robots and Systems (IROS)}. pp. 769--776.
\newblock \doi{10.1109/IROS.2017.8202237}.

\bibitem[{Lenz et~al.(2015)Lenz, Lee and Saxena}]{2015_deep}
Lenz I, Lee H and Saxena A (2015) Deep learning for detecting robotic grasps.
\newblock \emph{The International Journal of Robotics Research (IJRR)} 34(4-5):
  705--724.

\bibitem[{Li et~al.(2018)Li, Chen and Lee}]{2018_sonet}
Li J, Chen BM and Lee GH (2018) So-net: Self-organizing network for point cloud
  analysis.
\newblock In: \emph{Proceedings of the IEEE Conference on Computer Vision and
  Pattern Recognition (CVPR)}. pp. 9397--9406.

\bibitem[{Li et~al.(2021)Li, Kong, Chu, Li, Wang and Li}]{2021_simultaneous}
Li Y, Kong T, Chu R, Li Y, Wang P and Li L (2021) Simultaneous semantic and
  collision learning for 6-dof grasp pose estimation.
\newblock In: \emph{IEEE International Conference on Intelligent Robots and
  Systems (IROS)}. IEEE, pp. 3571--3578.

\bibitem[{Liang et~al.(2019)Liang, Ma, Li, Gorner, Tang, Fang, Sun and
  Zhang}]{2019_pointnetgpd}
Liang H, Ma X, Li S, Gorner M, Tang S, Fang B, Sun F and Zhang J (2019)
  Pointnetgpd: Detecting grasp configurations from point sets.
\newblock In: \emph{International Conference on Robotics and Automation
  (ICRA)}. IEEE, pp. 3629--3635.

\bibitem[{Lin et~al.(2017)Lin, Goyal, Girshick, He and
  Dollar}]{2017_focal_loss}
Lin TY, Goyal P, Girshick R, He K and Dollar P (2017) Focal loss for dense
  object detection.
\newblock In: \emph{Proceedings of the IEEE International Conference on
  Computer Vision (ICCV)}. pp. 2980--2988.

\bibitem[{Liu et~al.(2019)Liu, Tang, Lin and Han}]{2019_point_voxel}
Liu Z, Tang H, Lin Y and Han S (2019) Point-voxel cnn for efficient 3d deep
  learning.
\newblock In: \emph{Advances in Neural Information Processing Systems
  (NeurIPS)}, volume~32.

\bibitem[{Lou et~al.(2020)Lou, Yang and Choi}]{2020_reachability}
Lou X, Yang Y and Choi C (2020) Learning to generate 6-dof grasp poses with
  reachability awareness.
\newblock In: \emph{IEEE International Conference on Robotics and Automation
  (ICRA)}. pp. 1532--1538.

\bibitem[{Mahler et~al.(2017)Mahler, Liang, Niyaz, Laskey, Doan, Liu, Ojea and
  Goldberg}]{2017_dexnet2}
Mahler J, Liang J, Niyaz S, Laskey M, Doan R, Liu X, Ojea JA and Goldberg K
  (2017) Dex-net 2.0: Deep learning to plan robust grasps with synthetic point
  clouds and analytic grasp metrics.
\newblock In: \emph{Robotics: Science and Systems (RSS)}.

\bibitem[{Mahler et~al.(2016)Mahler, Pokorny, Hou, Roderick, Laskey, Aubry,
  Kohlhoff, Kroger, Kuffner and Goldberg}]{2016_dexnet1}
Mahler J, Pokorny FT, Hou B, Roderick M, Laskey M, Aubry M, Kohlhoff K, Kroger
  T, Kuffner J and Goldberg K (2016) Dex-net 1.0: A cloud-based network of 3d
  objects for robust grasp planning using a multi-armed bandit model with
  correlated rewards.
\newblock In: \emph{IEEE international conference on robotics and automation
  (ICRA)}. IEEE, pp. 1957--1964.

\bibitem[{Miller and Allen(1999)}]{fc_gws}
Miller A and Allen PK (1999) Examples of 3d grasp quality computations.
\newblock In: \emph{IEEE International Conference on Robotics and Automation
  (ICRA)}.

\bibitem[{Morrison et~al.(2018)Morrison, Corke and Leitner}]{2018_close}
Morrison D, Corke P and Leitner J (2018) Closing the loop for robotic grasping:
  A real-time, generative grasp synthesis approach.
\newblock In: \emph{Robotics: Science and Systems Conference (RSS)}.

\bibitem[{Morrison et~al.(2020)Morrison, Corke and Leitner}]{2020_egad}
Morrison D, Corke P and Leitner J (2020) Egad! an evolved grasping analysis
  dataset for diversity and reproducibility in robotic manipulation.
\newblock In: \emph{IEEE Robotics and Automation Letters (RAL)}, volume~5. pp.
  4368--4375.

\bibitem[{Mousavian et~al.(2019)Mousavian, Eppner and Fox}]{2019_6dof_graspnet}
Mousavian A, Eppner C and Fox D (2019) 6-dof graspnet: Variational grasp
  generation for object manipulation.
\newblock In: \emph{Proceedings of the IEEE/CVF International Conference on
  Computer Vision}. pp. 2901--2910.

\bibitem[{Murali et~al.(2020)Murali, Mousavian, Eppner, Paxton and
  Fox}]{2019_6dof_target_driven}
Murali A, Mousavian A, Eppner C, Paxton C and Fox D (2020) 6-dof grasping for
  target-driven object manipulation in clutter.
\newblock In: \emph{IEEE International Conference on Robotics and Automation
  (ICRA)}. pp. 6232--6238.

\bibitem[{Nguyen(1987)}]{force_closure2}
Nguyen VD (1987) Constructing force-closure grasps in 3d.
\newblock In: \emph{IEEE International Conference on Robotics and Automation
  (ICRA)}, volume~4. pp. 240--245.

\bibitem[{Ni et~al.(2020)Ni, Zhang, Zhu and Cao}]{2020_pointnet++_grasping}
Ni P, Zhang W, Zhu X and Cao Q (2020) Pointnet++ grasping: Learning an
  end-to-end spatial grasp generation algorithm from sparse point clouds.
\newblock \emph{IEEE International Conference on Robotics and Automation
  (ICRA)} .

\bibitem[{Ponce and Faverjon(1995)}]{1995_3finger}
Ponce J and Faverjon B (1995) On computing three-finger force-closure grasps of
  polygonal objects.
\newblock In: \emph{IEEE Transactions on robotics and automation}. p.
  868–881.

\bibitem[{Ponce et~al.(1997)Ponce, Sullivan, Sudsang, Boissonnat and
  Merlet}]{1997_4finger}
Ponce J, Sullivan S, Sudsang A, Boissonnat JD and Merlet JP (1997) On computing
  three-finger force-closure grasps of polygonal objects.
\newblock In: \emph{International Journal of Robotics Research (IJRR)}. pp.
  11--35.

\bibitem[{Qi et~al.(2019)Qi, Litany, He and Guibas}]{2019_hough_voting}
Qi CR, Litany O, He K and Guibas LJ (2019) Deep hough voting for 3d object
  detection in point clouds.
\newblock In: \emph{Proceedings of the IEEE International Conference on
  Computer Vision (ICCV)}. pp. 9277--9286.

\bibitem[{Qi et~al.(2017{\natexlab{a}})Qi, Su, Mo and Guibas}]{2017_pointnet}
Qi CR, Su H, Mo K and Guibas LJ (2017{\natexlab{a}}) Pointnet: Deep learning on
  point sets for 3d classification and segmentation.
\newblock In: \emph{Proceedings of the IEEE conference on computer vision and
  pattern recognition (CVPR)}. pp. 652--660.

\bibitem[{Qi et~al.(2017{\natexlab{b}})Qi, Yi, Su and Guibas}]{2017_pointnet++}
Qi CR, Yi L, Su H and Guibas LJ (2017{\natexlab{b}}) Pointnet++: Deep
  hierarchical feature learning on point sets in a metric space.
\newblock In: \emph{Advances in neural information processing systems}.

\bibitem[{Qin et~al.(2020)Qin, Chen, Zhu, Song, Xu and Su}]{2020_s4g}
Qin Y, Chen R, Zhu H, Song M, Xu J and Su H (2020) S4g: Amodal single-view
  single-shot se (3) grasp detection in cluttered scenes.
\newblock In: \emph{Conference on Robot Learning (CoRL)}. PMLR, pp. 53--65.

\bibitem[{Redmon and Angelova(2015)}]{2015_real_convolutional}
Redmon J and Angelova A (2015) Real-time grasp detection using convolutional
  neural networks : 1316--1322.

\bibitem[{Ren et~al.(2015)Ren, He, Girshick and Sun}]{2015faster_rcnn}
Ren S, He K, Girshick R and Sun J (2015) Faster r-cnn: Towards real-time object
  detection with region proposal networks.
\newblock In: \emph{Advances in neural information processing systems}. pp.
  91--99.

\bibitem[{Sahbani et~al.(2012)Sahbani, El-Khoury and Bidaud}]{grasp_overview}
Sahbani A, El-Khoury S and Bidaud P (2012) An overview of 3d object grasp
  synthesis algorithms.
\newblock In: \emph{Robotics and Autonomous Systems}, volume~60. pp. 326--336.

\bibitem[{Saxena et~al.(2008)Saxena, Driemeyer and Ng}]{2008_grasp_point}
Saxena A, Driemeyer J and Ng AY (2008) Robotic grasping of novel objects using
  vision.
\newblock \emph{The International Journal of Robotics Research} 27(2):
  157--173.
\newblock \doi{10.1177/0278364907087172}.
\newblock \urlprefix\url{https://doi.org/10.1177/0278364907087172}.

\bibitem[{Shao et~al.(2020)Shao, Ferreira, Jorda, Nambiar, Luo and
  Solowjow}]{2020_unigrasp}
Shao L, Ferreira F, Jorda M, Nambiar V, Luo J and Solowjow E (2020) Unigrasp:
  Learning a unified model to grasp with multifingered robotic hands.
\newblock In: \emph{IEEE Robotics and Automation Letters}, volume~5. IEEE, pp.
  2286--2293.

\bibitem[{Smith et~al.(1999)Smith, Lee, Goldberg, Bohringer and
  Craig}]{1999_parallel-jaw}
Smith G, Lee E, Goldberg K, Bohringer K and Craig J (1999) Computing
  parallel-jaw grips.
\newblock In: \emph{IEEE International Conference on Robotics and Automation
  (ICRA)}. p. 1897–1903.

\bibitem[{Sundermeyer et~al.(2021)Sundermeyer, Mousavian, Triebel and
  Fox}]{2021_contact_graspnet}
Sundermeyer M, Mousavian A, Triebel R and Fox D (2021) Contact-graspnet:
  Efficient 6-dof grasp generation in cluttered scenes.
\newblock \emph{IEEE International Conference on Robotics and Automation
  (ICRA)} : 13438--13444.

\bibitem[{ten Pas et~al.(2017)ten Pas, Gualtieri, Saenko and
  Platt}]{2017_gpd_ijrr}
ten Pas A, Gualtieri M, Saenko K and Platt R (2017) Grasp pose detection in
  point clouds.
\newblock \emph{The International Journal of Robotics Research (IJRR)}
  36(13-14): 1455--1473.
\newblock \doi{10.1177/0278364917735594}.
\newblock \urlprefix\url{https://doi.org/10.1177/0278364917735594}.

\bibitem[{Todorov et~al.(2012)Todorov, Erez and Tassa}]{2012_mujoco}
Todorov E, Erez T and Tassa Y (2012) Mujoco: A physics engine for model-based
  control.
\newblock In: \emph{IEEE/RSJ International Conference on Intelligent Robots and
  Systems}. pp. 5026--5033.

\bibitem[{Wang et~al.(2021)Wang, Fang, Gou, Fang, Gao and Lu}]{2021_graspness}
Wang C, Fang H, Gou M, Fang H, Gao J and Lu C (2021) Graspness discovery in
  clutters for fast and accurate grasp detection.
\newblock In: \emph{Proceedings of the IEEE International Conference on
  Computer Vision (CVPR)}. pp. 15964--15973.

\bibitem[{Wang et~al.(2019)Wang, Tseng, Li, Jiang, Guo, Danielczuk, Mahler,
  Ichnowski and Goldberg}]{2019_adversarial_grasp}
Wang D, Tseng D, Li P, Jiang Y, Guo M, Danielczuk M, Mahler J, Ichnowski J and
  Goldberg K (2019) Adversarial grasp objects.
\newblock In: \emph{IEEE International Conference on Automation Science and
  Engineering (CASE)}. pp. 241--248.

\bibitem[{Watson et~al.(2017)Watson, Hughes and Iida}]{2017_real_time_grasp}
Watson J, Hughes J and Iida F (2017) Real-world, real-time robotic grasping
  with convolutional neural networks.
\newblock In: Gao Y, Fallah S, Jin Y and Lekakou C (eds.) \emph{Towards
  Autonomous Robotic Systems}. Cham: Springer International Publishing, pp.
  617--626.

\bibitem[{Wei et~al.(2021)Wei, Luo, Li, Xu, Zhong, Li and Wang}]{2021_gpr}
Wei W, Luo Y, Li F, Xu G, Zhong J, Li W and Wang P (2021) Gpr: Grasp pose
  refinement network for cluttered scenes.
\newblock In: \emph{IEEE International Conference on Robotics and Automation
  (ICRA)}. pp. 4295--4302.

\bibitem[{Wu et~al.(2020)Wu, Chen, Cao, Zhang, Tai, Sun and
  Jia}]{2020_grasp_proposal_net}
Wu C, Chen J, Cao Q, Zhang J, Tai Y, Sun L and Jia K (2020) Grasp proposal
  networks: An end-to-end solution for visual learning of robotic grasps.
\newblock In: Larochelle H, Ranzato M, Hadsell R, Balcan M and Lin H (eds.)
  \emph{Advances in Neural Information Processing Systems}, volume~33. Curran
  Associates, Inc., pp. 13174--13184.
\newblock
  \urlprefix\url{https://proceedings.neurips.cc/paper/2020/file/994d1cad9132e48c993d58b492f71fc1-Paper.pdf}.

\bibitem[{Wu et~al.(2019)Wu, Qi and Fuxin}]{2019_pointconv}
Wu W, Qi Z and Fuxin L (2019) Pointconv: Deep convolutional networks on 3d
  point clouds.
\newblock In: \emph{Proceedings of the IEEE Conference on Computer Vision and
  Pattern Recognition (CVPR)}. pp. 9621--9630.

\bibitem[{Xu et~al.(2021{\natexlab{a}})Xu, Chu and Vela}]{2021_gknet}
Xu R, Chu FJ and Vela PA (2021{\natexlab{a}}) Gknet: Grasp keypoint network for
  grasp candidates detection.
\newblock \emph{The International Journal of Robotics Research} :
  02783649211069569\doi{10.1177/02783649211069569}.

\bibitem[{Xu et~al.(2021{\natexlab{b}})Xu, Qi, Agrawal and
  Song}]{2021_adagrasp}
Xu Z, Qi B, Agrawal S and Song S (2021{\natexlab{b}}) Adagrasp: Learning an
  adaptive gripper-aware grasping policy.
\newblock In: \emph{IEEE International Conference on Robotics and Automation
  (ICRA)}. pp. 4620--4626.

\bibitem[{Yan et~al.(2019)Yan, Khansari, Hsu, Gong, Bai, Pirk and
  Lee}]{2019_sim2real}
Yan X, Khansari M, Hsu J, Gong Y, Bai Y, Pirk S and Lee H (2019) Data-efficient
  learning for sim-to-real robotic grasping using deep point cloud prediction
  networks.
\newblock \doi{10.48550/ARXIV.1906.08989}.
\newblock \urlprefix\url{https://arxiv.org/abs/1906.08989}.

\bibitem[{Zhang et~al.(2019)Zhang, Lan, Bai, Zhou, Tian and Zheng}]{2019_roi}
Zhang H, Lan X, Bai S, Zhou X, Tian Z and Zheng N (2019) Roi-based robotic
  grasp detection for object overlapping scenes.
\newblock In: \emph{IEEE/RSJ International Conference on Intelligent Robots and
  Systems (IROS)}. IEEE, pp. 4768--4775.

\bibitem[{Zhang et~al.(2022)Zhang, Yang, Wang, Zhao, Lan, Ding and
  Zheng}]{2022_regrad}
Zhang H, Yang D, Wang H, Zhao B, Lan X, Ding J and Zheng N (2022) Regrad: A
  large-scale relational grasp dataset for safe and object-specic robotic
  grasping in clutter.
\newblock \emph{IEEE Robotics and Automation Letters} .

\bibitem[{Zhang et~al.(2021)Zhang, Zhou, Lan, Li, Tian and
  Zheng}]{2019_oriented_anchor}
Zhang H, Zhou X, Lan X, Li J, Tian Z and Zheng N (2021) A real-time robotic
  grasping approach with oriented anchor box.
\newblock \emph{IEEE Transactions on Systems, Man, and Cybernetics: Systems}
  51(5): 3014--3025.
\newblock \doi{10.1109/TSMC.2019.2917034}.

\bibitem[{Zhao et~al.(2021{\natexlab{a}})Zhao, Zhang, Lan, Wang, Tian and
  Zheng}]{2021_regnet}
Zhao B, Zhang H, Lan X, Wang H, Tian Z and Zheng N (2021{\natexlab{a}}) Regnet:
  Region-based grasp network for single-shot grasp detection in point clouds.
\newblock In: \emph{IEEE International Conference on Robotics and Automation
  (ICRA)}. pp. 13474--13480.

\bibitem[{Zhao et~al.(2021{\natexlab{b}})Zhao, Jiang, Jia, Torr and
  Koltun}]{2021_point_transformer}
Zhao H, Jiang L, Jia J, Torr P and Koltun V (2021{\natexlab{b}}) Point
  transformer.
\newblock In: \emph{Proceedings of the IEEE Conference on Computer Vision and
  Pattern Recognition (CVPR)}. pp. 16259--16268.

\bibitem[{Zheng and Qian(2005)}]{force_closure1}
Zheng Y and Qian W (2005) Coping with the grasping uncertainties in
  force-closure analysis.
\newblock In: \emph{International Journal of Robotics Research (IJRR)},
  volume~24. pp. 311--327.

\bibitem[{Zhou et~al.(2018)Zhou, Lan, Zhang, Tian, Zhang and
  Zheng}]{2018_anchor}
Zhou X, Lan X, Zhang H, Tian Z, Zhang Y and Zheng N (2018) Fully convolutional
  grasp detection network with oriented anchor box.
\newblock In: \emph{IEEE/RSJ International Conference on Intelligent Robots and
  Systems (IROS)}. IEEE, p. 7223–7230.

\bibitem[{Zhou and Tuzel(2018)}]{2018_voxelnet}
Zhou Y and Tuzel O (2018) Voxelnet: End-to-end learning for point cloud based
  3d object detections.
\newblock In: \emph{Proceedings of the IEEE Conference on Computer Vision and
  Pattern Recognition (CVPR)}. pp. 4490--4499.

\end{thebibliography}
\end{document}